\documentclass[10pt,journal,compsoc]{IEEEtran}
\usepackage{cite}
\usepackage{multirow}
\usepackage{dcolumn}
\usepackage{amsmath}
\usepackage{kantlipsum}
\usepackage{url}
\usepackage{verbatim}
\usepackage{colortbl}
\usepackage{amssymb}
\usepackage{arydshln}
\usepackage{multirow}
\usepackage[citecolor=red,linkcolor=blue,colorlinks=true,pdftex,plainpages=false]{hyperref}
\usepackage{microtype}
\newcolumntype{K}[1]{>{\centering\arraybackslash}p{#1}}
\usepackage{booktabs}
\usepackage[flushleft]{threeparttable}
\usepackage{hhline}
\usepackage{wrapfig}
\usepackage{algorithm}
\usepackage{algorithmic}
\usepackage{mathtools}
\usepackage{placeins}
\usepackage{kantlipsum}
\usepackage{dsfont}
\usepackage[export]{adjustbox}
\usepackage{wrapfig}
\usepackage{amsmath}

\newcommand{\RNum}[1]{\uppercase\expandafter{\romannumeral #1\relax}}
\ifCLASSOPTIONcompsoc
\usepackage[caption=false, font=normalsize, labelfont=sf, textfont=sf]{subfig}
\else
\usepackage[caption=false, font=footnotesize]{subfig}
\usepackage{subcaption}
\include[outdir=./(folder)/]{epstopdf}
\usepackage[pdf]{pstricks}
\usepackage{subfig}
\usepackage[utf8]{inputenc}
\usepackage{bbm}
\usepackage{tikz}
\fi
\usepackage{tikz} 
\usetikzlibrary{fit}
\usetikzlibrary{shapes}
\newcommand{\enumber}[1]{%
    \begin{tikzpicture}[remember picture]
        \node[inner sep=0pt](a){#1};
    \end{tikzpicture}%
    \begin{tikzpicture}[overlay, remember picture]
        \node[draw, red, fit=(a), ellipse, inner sep=1pt, line width=1.5pt]{};
    \end{tikzpicture}%
    }
\usepackage{acronym}

\newlength{\aligntop}
\newlength{\alignbot}
 

\begin{document}
\title{DISPATCH: \textbf{D}es\textbf{i}gn \textbf{S}pace Ex\textbf{p}lor\textbf{at}ion of \textbf{C}yber-P\textbf{h}ysical Systems}

\author{Prerit~Terway,~Kenza~Hamidouche,~and~Niraj~K.~Jha,~\IEEEmembership{Fellow,~IEEE}
\thanks{
Prerit Terway and Niraj K. Jha are with the Department of Electrical Engineering, and 
Kenza Hamidouche with the Department of Operations Research and Financial Engineering, Princeton 
University, Princeton, NJ, 08544 USA, e-mail:\{pterway, jha, kenzah\}@princeton.edu.}}

\IEEEtitleabstractindextext{%
\begin{abstract}
Design of cyber-physical systems (CPSs) is a challenging task that involves searching over a large search space of various CPS configurations and possible values of components composing the system. Hence, there is a need for sample-efficient CPS design space exploration to select the system architecture and component values that meet the target system requirements. We address this challenge by formulating CPS design as a multi-objective optimization problem and propose DISPATCH, a two-step methodology for sample-efficient search over the design space. First, we use a genetic algorithm to search over discrete choices of system component values for architecture search and component selection or only component selection and terminate the algorithm even before meeting the system requirements, thus yielding a coarse design. In the second step, we use an inverse design to search over a continuous space to fine-tune the component values and meet the diverse set of system requirements. We use a neural network as a surrogate function for the inverse design of the system. The neural network, converted into a mixed-integer linear program, is used for active learning to sample component values efficiently in a continuous search space. We illustrate the efficacy of DISPATCH on electrical circuit benchmarks: two-stage and three-stage transimpedence amplifiers. Simulation results show that the proposed methodology improves sample efficiency by 5-14$\times$ compared to a prior synthesis method that relies on reinforcement learning. It also synthesizes circuits with the best performance (highest bandwidth/lowest area) compared to designs synthesized using reinforcement learning, Bayesian optimization, or humans.
\end{abstract}

}
\maketitle

\IEEEdisplaynontitleabstractindextext

\IEEEpeerreviewmaketitle

\IEEEraisesectionheading{\section{Introduction}\label{sect:introduction}}
Cyber-physical systems (CPSs) form the foundation of various applications that include healthcare, 
smart grid, transportation, and smart home \cite{sampigethaya2013aviation, 
akmandor2017keep}. These systems consist of many interacting digital, analog, physical, and human 
components designed to perform specific functions through integrated physics and logic. 
CPS designs can be quite complex especially when they involve multidisciplinary 
analysis like optimizing the operating cost and range of an aircraft. For instance, conceptual 
designs account for around $80\%$ of the development cost in aircraft manufacturing 
\cite{tan2017comparison}. Hence, it is important to develop search techniques that minimize the 
number of costly CPS simulations and efficiently sample the design search space 
\cite{neema2014design}. 

CPS design entails finding both the system architecture as well as the appropriate component values 
with the goal of obtaining a final design that meets predefined system specifications. For example, a 
drone can have multiple architectures (quadcoptor, pentacopter, hexacopter) and each architecture 
requires selecting geometry, motor position, and orientation \cite{du2016computational}. Classical 
techniques for CPS design rely on humans for architecture selection \cite{lynch2016ontology}. The 
architecture is simulated over a range of component values to select the combination that meets the 
system requirements. Such a CPS design process limits the search space and 
costly CPS simulations increase development time.  

The development of more efficient design space exploration techniques must rely on a combination of
novel automated search techniques with the flexibility of incorporating existing knowledge from 
architectures designed by humans. CPS design can be formulated as a multi-objective optimization 
(MOO) problem to capture different system design objectives constrained by the components available 
for synthesis. For instance, when designing a drone, the constraints arise from the maximum torque the 
motor can produce with the objective of maximizing the payload and the distance the drone can travel.

To address the CPS design problem, we propose DISPATCH, a two-step design methodology that solves the 
problem in a sample-efficient manner.  The first step of DISPATCH either explores architectures 
as well as component values or only component values for a fixed architecture. 
We use a genetic algorithm (GA) to harness its benefits for exploration in a discrete 
design space and for the ease of encoding architectures with it.  We terminate the 
GA even before meeting the design requirements since GA is sample-inefficient and requires a
large number of iterations to meet the requirements. We thus obtain a coarse design in the first step.

In the second step, we fine-tune the coarse design to search for component values in a continuous 
search space. This step uses a neural network (NN) as a surrogate function to model system response.
We convert the NN into a mixed-integer linear program (MILP) to incorporate the constraints imposed 
by the inputs (i.e., component values), outputs (i.e., desired response), and the NN. We use this 
formulation to obtain an \emph{inverse design} of the system to find the component values that 
satisfy a set of system constraints.  A feasible solution of the MILP yields the component values 
that are used to simulate the system. Otherwise, we simulate the system with a random combination of 
permissible component values. The NN thus enables \textit{active learning} through generation of 
input samples until the system requirements are satisfied or the sampling budget is exhausted. 
The MILP solution to the NN acts like an \emph{acquisition function}.

We summarize the major contributions of this article as follows:
\begin{itemize}
\item We formulate CPS design as an MOO problem and propose DISPATCH, a two-step method for 
sample-efficient CPS synthesis.
\item In the first step of DISPATCH, we explore a large discrete search space using GA and synthesize 
a coarse design.
\item In the second step of DISPATCH, we explore a continuous search space of component values using 
an NN as a surrogate function. We use an MILP formulation of the NN to obtain an \emph{inverse design} 
of the system.
\item We demonstrate that DISPATCH requires much fewer simulations to synthesize valid designs 
compared to methodologies that are based on reinforcement learning (RL), Bayesian optimization, 
or human design. 
\end{itemize}

The rest of the article is organized as follows. In Section \ref{sect:related work}, we 
discuss related work. Section \ref{sect:background} provides the necessary background, followed by 
a simple motivational example in Section \ref{sec:motivation}. In Section \ref{sect:methodology}, we 
introduce the methodology for sample-efficient CPS design. We apply the methodology to the synthesis
of electrical circuits in Section \ref{sect:results}.  Finally, Section \ref{sec:conclusion} 
concludes the article. 

\section{Related work}
\label{sect:related work}
In this section, we review past work on solving MOO problems and automated system synthesis. For the 
latter, we review existing design techniques for analog circuits that we use later as
benchmarks to validate our methodology. Like CPS, the design of analog circuits requires searching 
over a large space of possible architectures and component values with the goal of optimizing specific 
circuit objectives. 

\subsection{Search techniques}
\label{subsec:relatedWorkSearch}
CPS design often has multiple objectives that need to be optimized based on system requirements 
\cite{cao2017distributed}. GA, a type of evolutionary algorithm (EA), is known to be
suitable for addressing MOO problems \cite{zebulum2018evolutionary}. A seminal work 
in this field is the NSGA-\RNum{2} algorithm \cite{996017} that uses a fast non-dominated sorting 
approach to reduce computational complexity by an order of magnitude.
EAs have several variants, e.g., differential evolution that perturbs 
randomly selected population members based on the difference between selected individuals 
\cite{kenneth1999price}, swarm intelligence that includes ant colony optimization \cite{4129846}, 
and particle swarm optimization \cite{kennedy1995particle}.
Genetic programming \cite{banzhaf1998genetic} is another 
evolutionary approach that evolves computer programs represented as tree structures.  Though EAs are 
good at exploration, lack of gradient information during search makes them sample-inefficient.

Gradient-based search, such as RL, can address the sample-inefficiency problem of EAs 
\cite{wang2018learning} at the cost of poorer exploration.  RL is used for system design by learning a 
policy to obtain component values based on the current system state \cite{powell2019unified, 
sutton2018reinforcement}. 
The recent spurt in interest involving the use of RL combined with deep 
learning is due to \cite{mnih2013playing}. This approach is based on training a convolutional NN to 
learn a policy for playing Atari games. This work was extended to the continuous action space 
in \cite{lillicrap2015continuous} to solve simulated physics tasks and learn end-to-end policies 
that are more sample-efficient than those discussed in \cite{mnih2013playing}. 

\subsection{System synthesis}
\label{sunsec:relatedWorkSystemSynthesis}

EA was widely used in the 1990's and early 2000's for synthesis of electrical circuits 
\cite{zebulum2018evolutionary}.  The use of EA to obtain the topology of analog circuits was
pioneered in \cite{koza1997automated}.  Parallel GA and circuit-construction primitives 
were used to create circuit graphs to evolve designs for an analog filter and amplifier
in \cite{lohn1999circuit}.  Similarly, EA and simulated annealing were combined for analog circuit 
synthesis in \cite{alpaydin2003evolutionary}. 
In \cite{passos2019two}, surrogate modeling with 
EA was proposed to synthesize a low-noise amplifier. 
Synthesis of amplifiers and filters was
done by solving constrained MOO with NSGA-\RNum{2} in \cite{nicosia2007evolutionary}. 

More recent works have primarily focused on determining component values for a fixed architecture 
in a sample-efficient manner. The method proposed in \cite{wang2018learning} uses deep deterministic 
policy gradient (DDPG), a form of RL, for sample-efficient synthesis of two-stage and three-stage 
transimpedence amplifiers, for this purpose.  The method in \cite{settaluri2020autockt} 
proposes deep RL combined with transfer learning over a sparse design space to synthesize analog 
circuits. The method in \cite{lyu2018batch} uses Bayesian optimization with an ensemble of 
acquisition functions to tackle complex mathematical functions and synthesize electrical circuits. 

The drawbacks of existing search techniques can be summarized as follows:
\begin{itemize}
    \item GA is sample-inefficient and often needs to repeat costly simulations multiple times to 
obtain an acceptable design.
    \item The CPS design problem is formulated as a weighted sum of multiple objectives. In the 
optimization process, the weights are determined using domain expertise or through multiple 
simulations, making the design procedure inefficient.
    \item Most CPS design formulations assume that the architecture is fixed and only focus on 
selecting component values, thereby limiting the search space and possibly missing out on novel
designs.
    \item Bayesian optimization based design techniques require a large amount of time to select the 
next sampling point for simulation \cite{wang2018learning}, thereby making the optimization process 
very slow.
\end{itemize}

\section{Background}
\label{sect:background}
Fast CPS synthesis warrants exploration in a sample-efficient manner. To this end, we formulate CPS 
design as an MOO problem and propose DISPATCH, a two-step CPS design methodology based on GA and 
\emph{inverse design}. Next, we provide background on MOO formulation of CPS design problems,
GA, and inverse design.

\subsection{CPS design through multi-objective optimization}
\label{multi-objective}
When designing CPS, the designer is interested in achieving multiple system objectives under
various constraints. This can be done by formulating CPS design as an MOO problem.  The solutions to 
the MOO problem indicate a set of choices that enable the best tradeoffs among competing objectives. 
These solutions constitute a \emph{non-dominated} set and lie on a surface called the \emph{Pareto 
front} \cite{van1998evolutionary}. When some decision variables can only 
take integer values, and the  constraints and objectives are linear, the problem can be
formulated as an MILP. Many CPS designs fall into this category. For instance, in the design of 
multicopters, the number of motors is an integer and the constraint on the torque produced by the 
motor is linear \cite{du2016computational}. The multicopter design objective may be to maximize the 
payload and distance it can travel.  More formally, the problem of CPS design can be formulated as 
follows:

\begin{equation}
\label{eq: MultiObjEquGeneric}
    \begin{aligned}
    & \underset{\boldsymbol{x}, \boldsymbol{y}}{\text{minimize}}
    & & f_{m}(\boldsymbol{x}, \boldsymbol{y}), \ m=1,2, \ldots, M \\
    & \text{subject to}
    & & g_{j}(\boldsymbol{x}, \boldsymbol{y}) \geq 0, j=1,2, \ldots, J \\
    &&& h_{k}(\boldsymbol{x}, \boldsymbol{y})=0, k=1,2, \ldots, K \\
    &&& x_{i}^{L} \leq x_{i} \leq x_{i}^{U}, i=1,2, \ldots, n
    \end{aligned}
\end{equation}
\noindent
where $\boldsymbol{y}$ represents the search space over all possible architectures, 
$\boldsymbol{x}$ is the search space over all possible component values, and $f_{m}(\boldsymbol{x}, \boldsymbol{y})$ is the $m^{th}$ objective.
The system must also satisfy $J$ inequality constraints given by $g_{j}(\boldsymbol{x}, \boldsymbol{y})$ and 
$K$ equality constraints given by $h_{k}(\boldsymbol{x}, \boldsymbol{y})$.
As an example, for 
electrical circuits, the constraints may be on the maximum power the system can consume or the minimum 
bandwidth that it must achieve.  The lower bound for the value $x_{i}$ of component $i$ is $x_{i}^{L}$ and 
the upper bound is $x_{i}^{U}$.  
Next, we introduce GA and NN-based inverse system design.

\subsection{Genetic algorithm}
\label{GA}
CPS designs often rely on human expertise.  However, this may lead to failure to  
find an architecture that meets all system requirements or require long simulation runs to explore 
the large design space.  In the first step of DISPATCH, we harness GA's exploration capability and 
ease of encoding an architecture~\cite{goldberg1988genetic} to solve the MOO formulation of CPS design.
GA evolves a population of individual solutions through multiple generations.  The main steps of GA 
are as follows:

\begin{enumerate}
    \item Each individual in the first generation of a population is represented with a 
\textit{chromosome}. A chromosome is a sequence of genes.
    \item Each individual is evaluated on the basis of how well it meets its multiple objectives. 
    \item A subset of these individuals is selected to produce children for the next generation.
    \item Pairs of randomly chosen individuals (parents) from this subset undergo
reproduction using \emph{crossover} by combining the genes of the parents to create children. 
    \item The genes of each child are \emph{mutated} by perturbing them with low probability to 
facilitate exploration across its various dimensions. 
    \item The best performing individuals in the current generation and the children (after
Step 5) are retained. 
    \item Steps 2-6 are repeated until one of the stopping criteria is met.
\end{enumerate}

\begin{figure}[!ht]
    \includegraphics[scale=0.33]{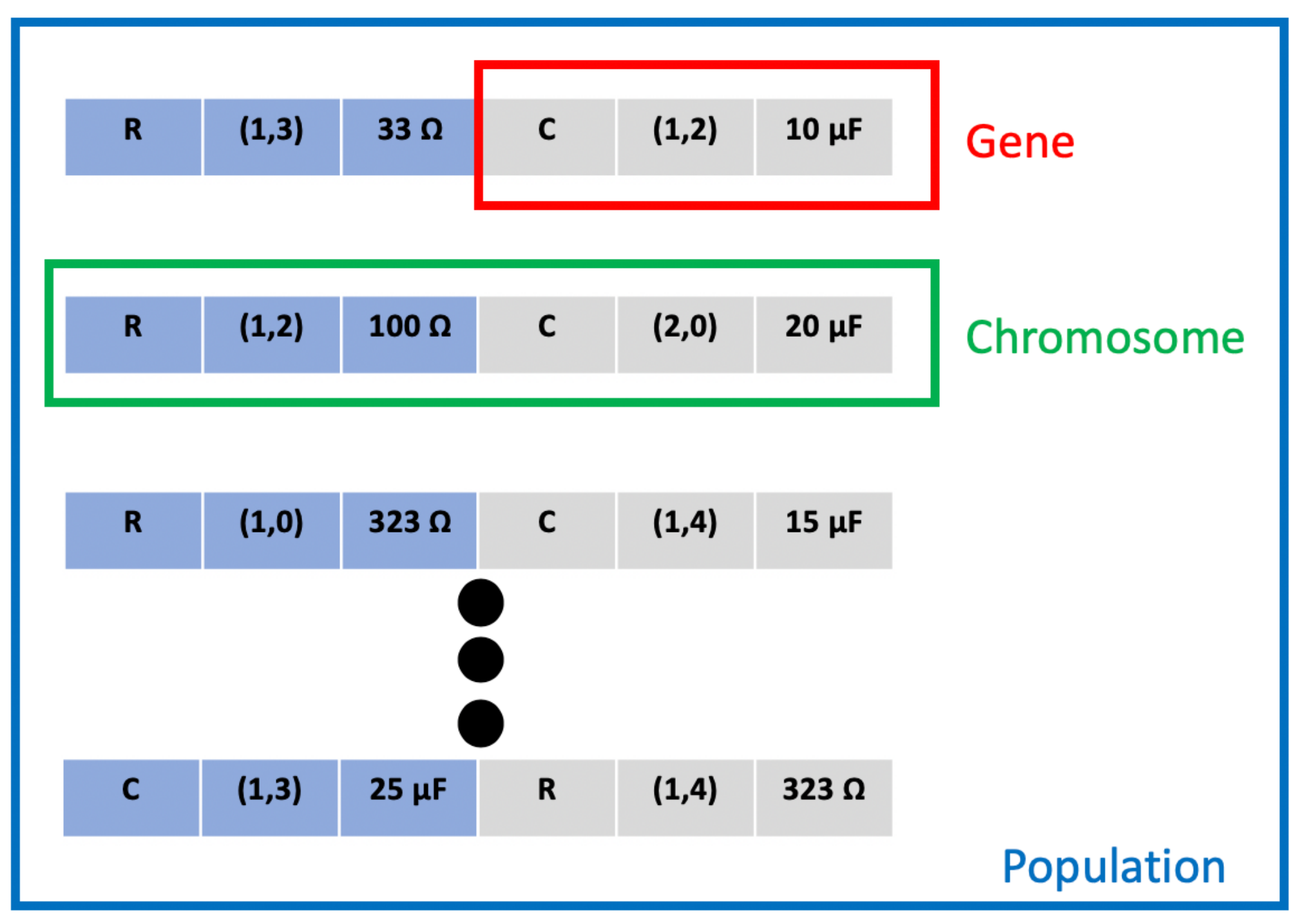}
    \centering
    \caption{Details of different elements of GA. Each component is color-coded 
(resistor in blue and capacitor in grey). Gene depicted in a red box, chromosome in a green box, and 
population in a blue box.}
\label{fig:GA details}
\end{figure}

In our case, we use GA to evolve CPS architectures. A \emph{chromosome} corresponds to 
an architecture. \emph{Genes} encode details of a particular component, like its type, connecting 
nodes, and value.  We evaluate the architecture through a CPS simulator. We use tournament selection 
to select the individuals that undergo reproduction \cite{blickle1996comparison}. 
Fig.~\ref{fig:GA details} shows a population of individuals, i.e., chromosomes, and genes. Here, the
system architecture corresponds to an electrical circuit.  GA requires a large number of simulations 
to obtain a design that meets all specifications, thereby pointing to the need for a 
sample-efficient design procedure.

\subsection{Inverse system design using NN}
\label{MILP Formulation}
We overcome the sample inefficiency of GA by terminating it at an intermediate stage (before 
necessarily reaching a valid or acceptable design) and then using a sample-efficient gradient-based 
search to meet system requirements for a fixed architecture.  Akintunde et 
al.~\cite{akintunde2018reachability} proposed an MILP formulation of an NN in the context of neural 
agent-environment systems to solve the \emph{reachability} problem for an NN-based policy trained 
using RL.  Reachability indicates whether the NN can output the desired values using permissible inputs.
An NN is converted into an MILP by representing all hidden neurons with constraints defined 
as follows:

\begin{equation}
\label{eq:MILPConversion}
  \begin{aligned} C_{i}=\left\{\bar{x}_{j}^{(i)}\right.& \geq W_{j}^{(i)}
\bar{x}^{(i-1)}+b_{j}^{(i)}, \\ \bar{x}_{j}^{(i)} & \leq W_{j}^{(i)}
\bar{x}^{(i-1)}+b_{j}^{(i)}+Q \bar{\delta}_{j}^{(i)}, \\ \bar{x}_{j}^{(i)} &\left.\geq 0,
\bar{x}_{j}^{(i)} \leq Q\left(1-\bar{\delta}_{j}^{(i)}\right), j=1, \ldots , L^{(i)} \right\}. \\  \end{aligned}  
\end{equation}

\noindent
In Eq.~(\ref{eq:MILPConversion}), $\forall{i,j}$, $\bar{x}_{j}^{(i)}$ corresponds to the $j^{th}$ 
neuron in the $i^{th}$ layer, $L^{(i)}$ is the number of neurons in the $i^{th}$ layer, $W_{j}^{(i)}$ represents weights that determine the input to
$\bar{x}_{j}^{(i)}$, $\bar{x}^{(i-1)}$ represents outputs from the $(i-1)^{th}$ layer, 
${b}_{j}^{(i)}$ is the bias for neuron $\bar{x}_{j}^{(i)}$, $Q$ is larger than the largest possible 
magnitude of $W_{j}^{(i)} \bar{x}^{(i-1)}+b_{j}^{(i)}$, and $\bar{\delta}_{j}^{(i)}$ is defined as 
follows:
\begin{equation}
\label{eq:delta}
    \begin{aligned}
   \bar{\delta}_{j}^{(i)} \triangleq\left\{\begin{array}{l}0 \text { if } \bar{x}_{j}^{(m)}>0 \\ 1 \text { otherwise }\end{array}\right. 
   \end{aligned} 
\end{equation}
 The constraints imposed by the hidden neurons of the network are
obtained from the union of all the constraints ($C_{i}$) shown in Eq.~(\ref{eq:MILPConversion}).

This formulation is used in a system called CNMA\footnote{CNMA stands for Constrained optimization 
with Neural networks, MILP and Active learning.} \cite{CNMA} to obtain the input sample points while 
designing a system. If the solution to the MILP problem is feasible, the input that satisfies the 
constraints is used as the next sample point to simulate the CPS, otherwise the CPS is simulated 
using a random sample.  The NN is trained to predict the output corresponding to the inputs by 
minimizing the mean-squared error for all the simulated points.

Next, we illustrate the design of a simple electrical circuit as a motivational example 
to show how DISPATCH deals with architecture and component selection, which are
generic CPS design problems.

\section{Motivation}
\label{sec:motivation}
\begin{figure}[!h]
    \includegraphics[scale=0.35]{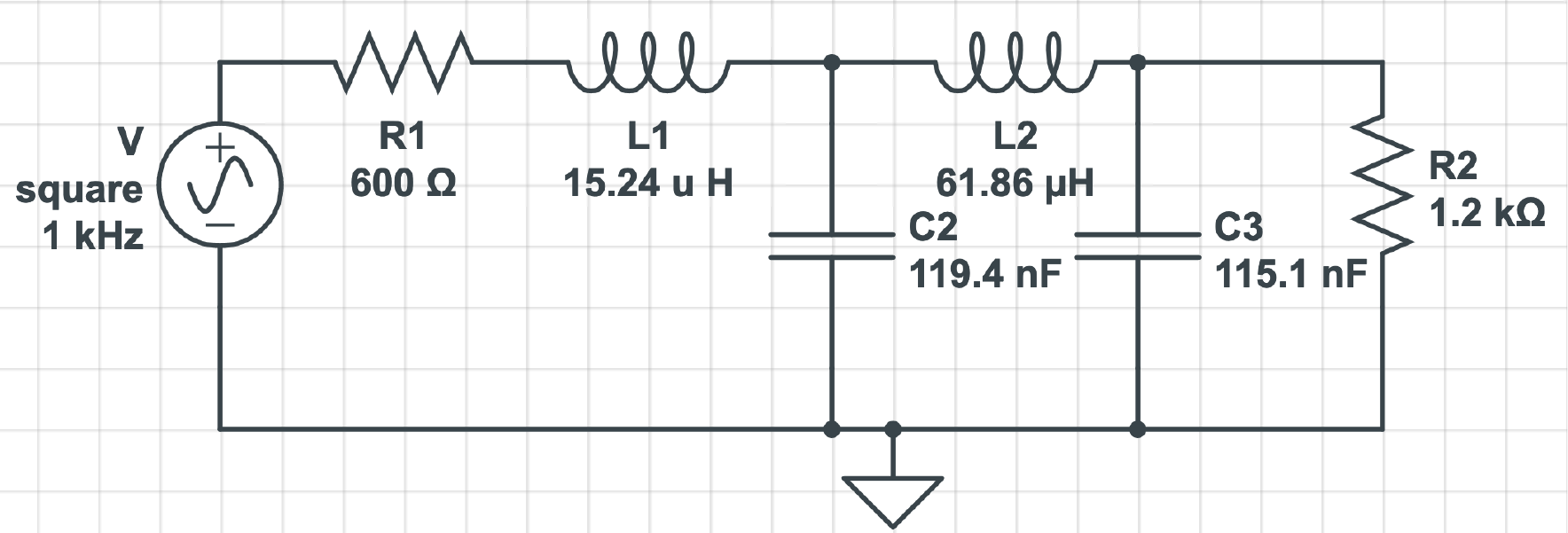}
    \centering
    \caption{Low pass Butterworth filter architecture and component values used as seed design.}
\label{fig:ButterworthFilter}
\end{figure}

We propose DISPATCH as a solution to the CPS design problem. Next, we illustrate its working 
through the design of a low-pass filter.
The objective is to obtain a \textit{unity gain ($0\  \rm dB$)} and a \textit{bandwidth of $1~\rm kHz$} 
while \textit{minimizing} the number of circuit components. The following discrete components are 
available:
\begin{itemize}
    \item Resistors: [1, 10, 600, 1200] $\Omega$
    \item Capacitance: [1e-12, 119.37e-9, 155.12e-9, 1e-5] F
    \item Inductance: [1e-6, 15.24e-3, 61.86e-3, 1e-2] H
\end{itemize}
The seed design is a Butterworth low-pass filter adapted from \cite{venturini_2015} and shown in 
Fig.~\ref{fig:ButterworthFilter}. We evolve it through GA till the point it gets close to meeting
the specifications.  We use a maximum of five nodes and 10 components in the circuit
architecture to expand the search space even though it is known that a low-pass filter can be
designed with just three nodes and two components (besides the supply).
The total number of choices for a gene is given by $(\# \textit{component types})\times
(\# \textit{connecting points})\times(\# \textit{different component values}) = 
3\times5^2\times4 = 300$ (each component has two connecting points resulting in $5^2$ connecting points) ~\cite{zebulum2018evolutionary}. In our example, since there may be 10 
components in the circuit, the search space size is $300^{10}$. Hence, an efficient strategy is 
required to search over such a large design space.

\begin{figure}[!ht]
    \includegraphics[scale=0.28]{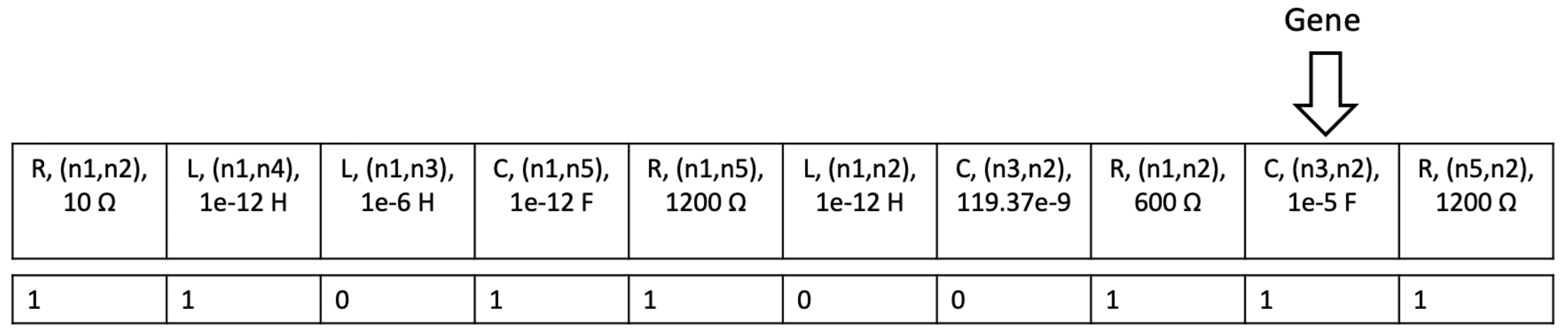}
    \centering
    \caption{Chromosome representation of a low-pass filter. The top row represents details of all 
the 10 components. It shows the component type given by (R, L, C) and its connecting nodes 
($n$1, $n$2, $n$3, \ldots) and value.  The bottom row indicates whether the component is active (1) 
or inactive (0) in the circuit.}
\label{fig:LowPassGene}
\end{figure}

Fig.~\ref{fig:LowPassGene} shows the chromosome representation of a circuit. A gene encodes the 
details of a component, including its type, connecting nodes, and value (shown in the top row). The 
bottom row shows whether the component is active (1) or inactive (0)~\cite{zebulum2018evolutionary}. 
We define the following objectives to find the coarse design by searching the architecture and 
component space using GA:
\begin{enumerate}
    \item Weighted sum of the difference between the desired \textit{magnitude} response of a 
first-order low-pass filter and the observed response. We set the weights in the \textit{passband} 
to $40$ and $1$ in the \textit{stopband} to give more importance to the response in 
the \textit{passband}.  Note that since GA synthesizes a coarse design, other values may work too. We use weighted sum here to cover the entire frequency range. 
    \item Weighted sum of the difference between the desired \textit{phase} response of a first-order 
low-pass filter and the observed response. The weights are the same as above.
    \item Number of \textit{active} components: a lower value implies fewer components.
\end{enumerate}

\begin{figure*}[htbp]
\centering
\subfloat[]{\includegraphics[height=1.2in, width=2.in]{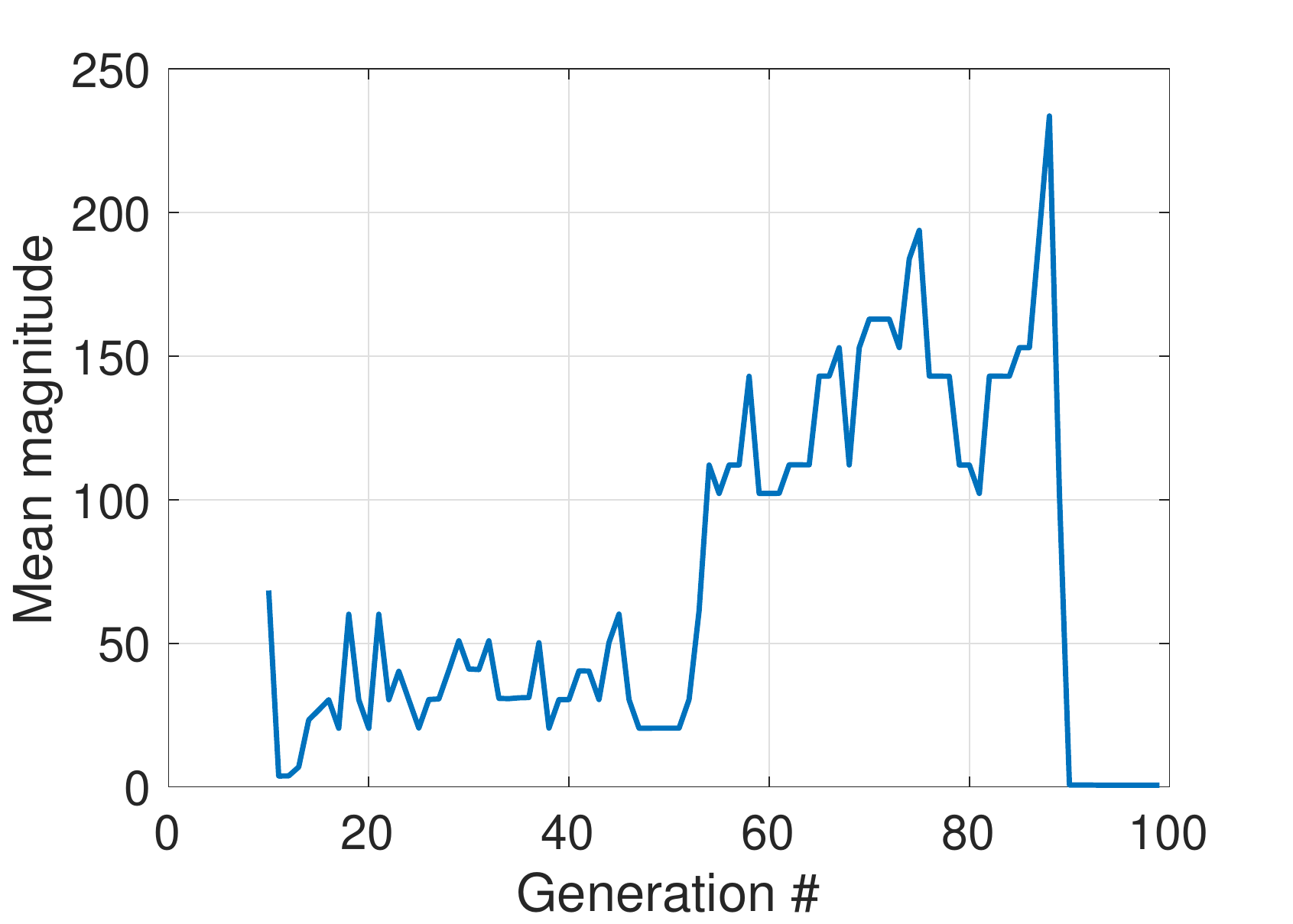}}
\subfloat[]{\includegraphics[height=1.2in, width=2.in]{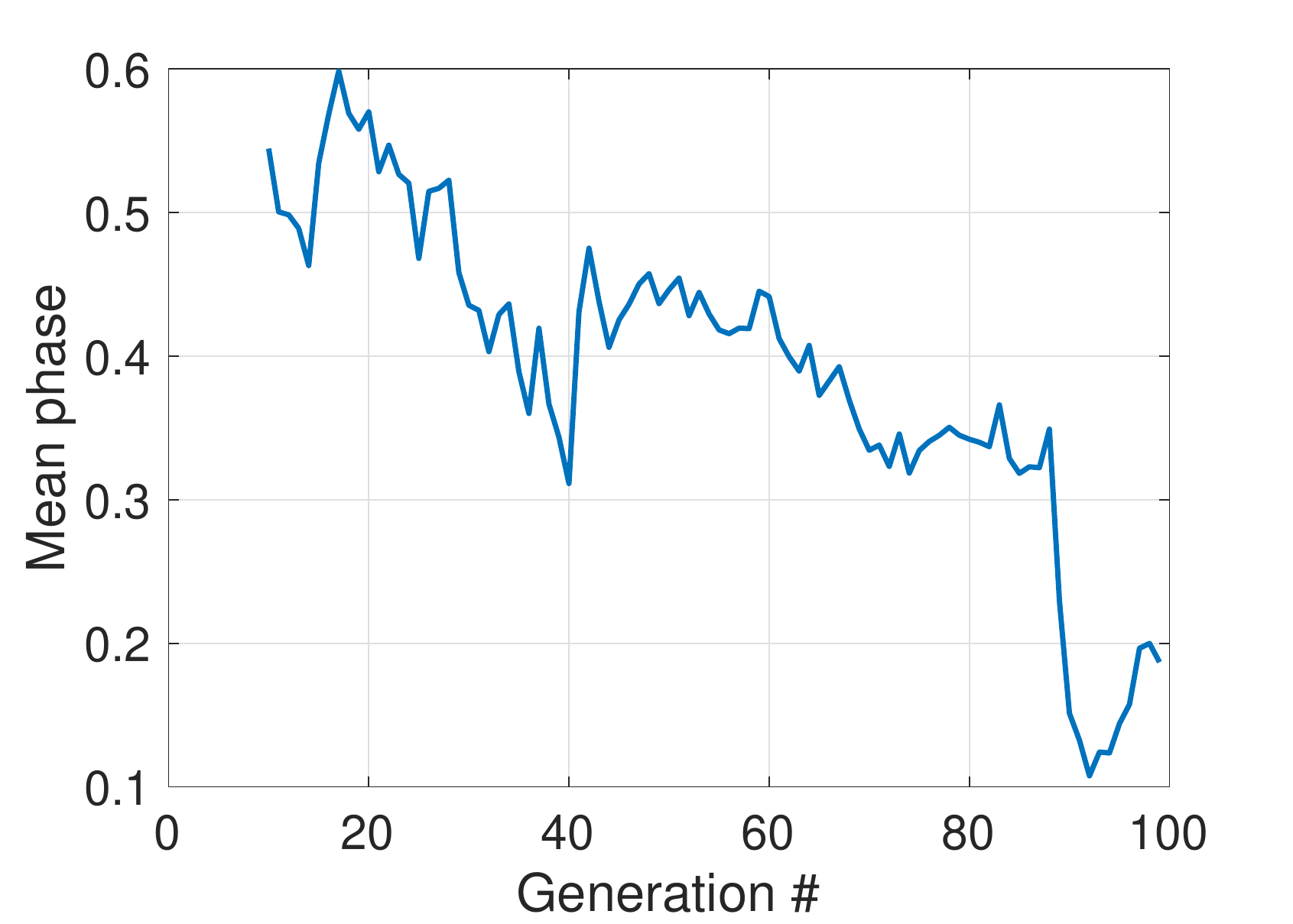}}
\subfloat[]{\includegraphics[height=1.2in, width=2.in]{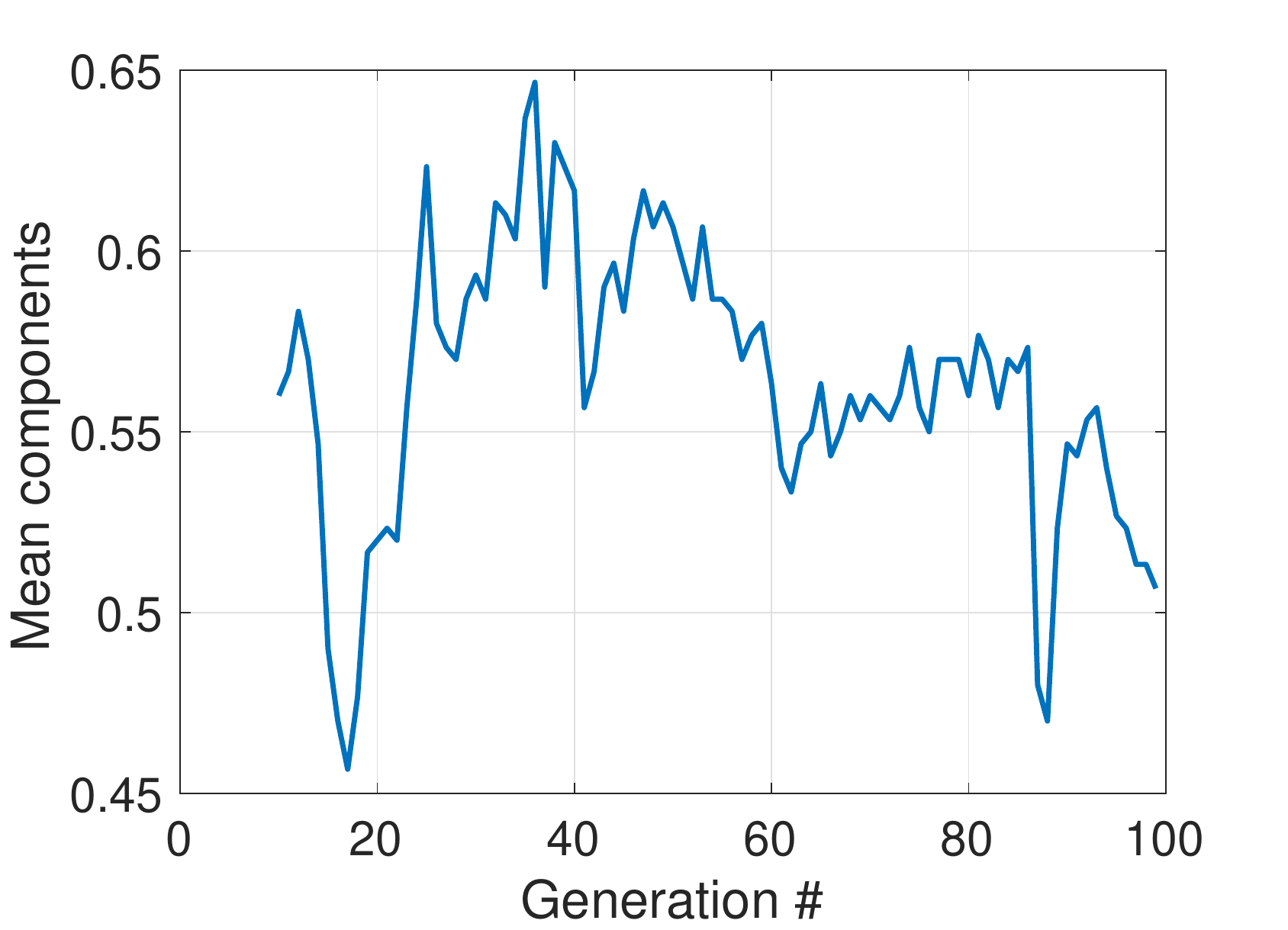}}
\caption{Scaled mean objectives across all individuals in a generation for (a) magnitude, (b)
phase, and (c) component values from the $10^{th}$ generation onwards for architecture search 
for low-pass filter design.}
\label{fig:GAEvolutionLowPass}
\end{figure*}

\begin{figure}[!ht]
    \includegraphics[scale=.50]{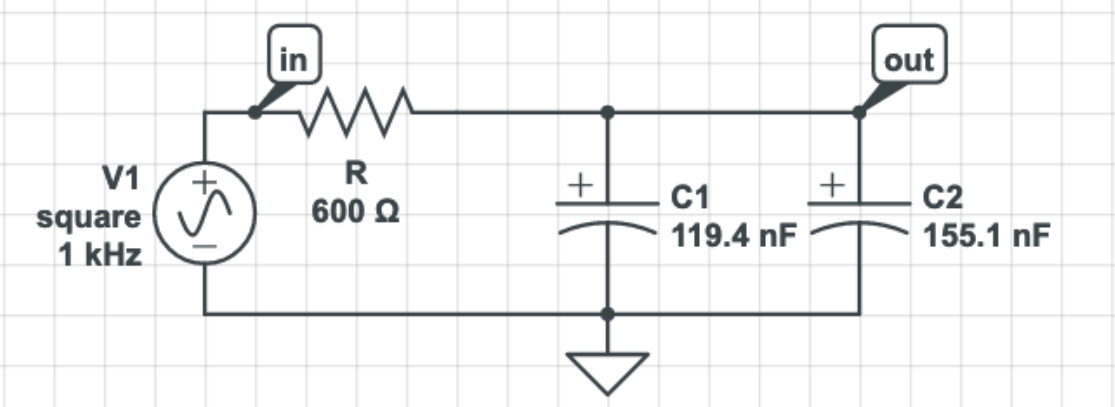}
    \centering
    \caption{Low-pass filter architecture evolved by GA. }
\label{fig:GASynthLowPass}
\end{figure}

We use GA for architecture search using 50 individuals comprising the seed design and other randomly 
generated individuals evolved over 100 generations. 
Fig.~\ref{fig:GAEvolutionLowPass} shows the scaled mean (with respect to the seed design) of 
the three objectives for the entire population from the $10^{th}$ generation onwards. We plot
the mean objective values after the $10^{th}$ generation because the initial GA generations have very 
high objective values.  The $x$-axis shows the generation number and the $y$-axis shows the mean value 
for the three objectives.  Since the mean values are scaled, the $y$-axis has no unit.     
There is a trade-off among the three objectives across generations, as evident from the figure. 
After 100 generations, we select a non-dominated individual from the Pareto front based on the first 
two objectives (magnitude and phase).  This individual (circuit) has a bandwidth of 966 Hz, phase 
of -46 degree, gain of 0 dB, and a total of three components. Since this is a coarse design, it does 
not meet the requirements yet. Fig.~\ref{fig:GASynthLowPass} shows the GA-evolved circuit. The circuit 
synthesized by GA can be simplified using human intervention by replacing the two parallel capacitors 
with a single equivalent capacitor, thus yielding a standard low-pass filter.  The next step is to 
fine-tune the component values in a continuous search space using a gradient-based search technique.

\begin{figure}[h]
    \includegraphics[scale=0.32]{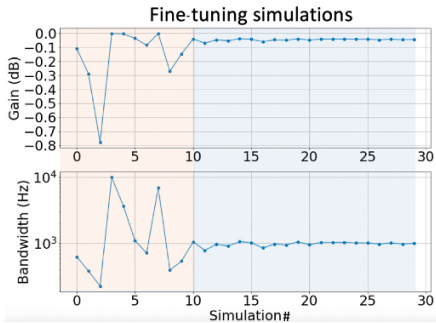}
    \centering
    \caption{Gain and bandwidth during fine-tuning using gradient-based search.}
\label{fig:CNMASimulation}
\end{figure}

\begin{figure}[h]
    \includegraphics[scale=0.2]{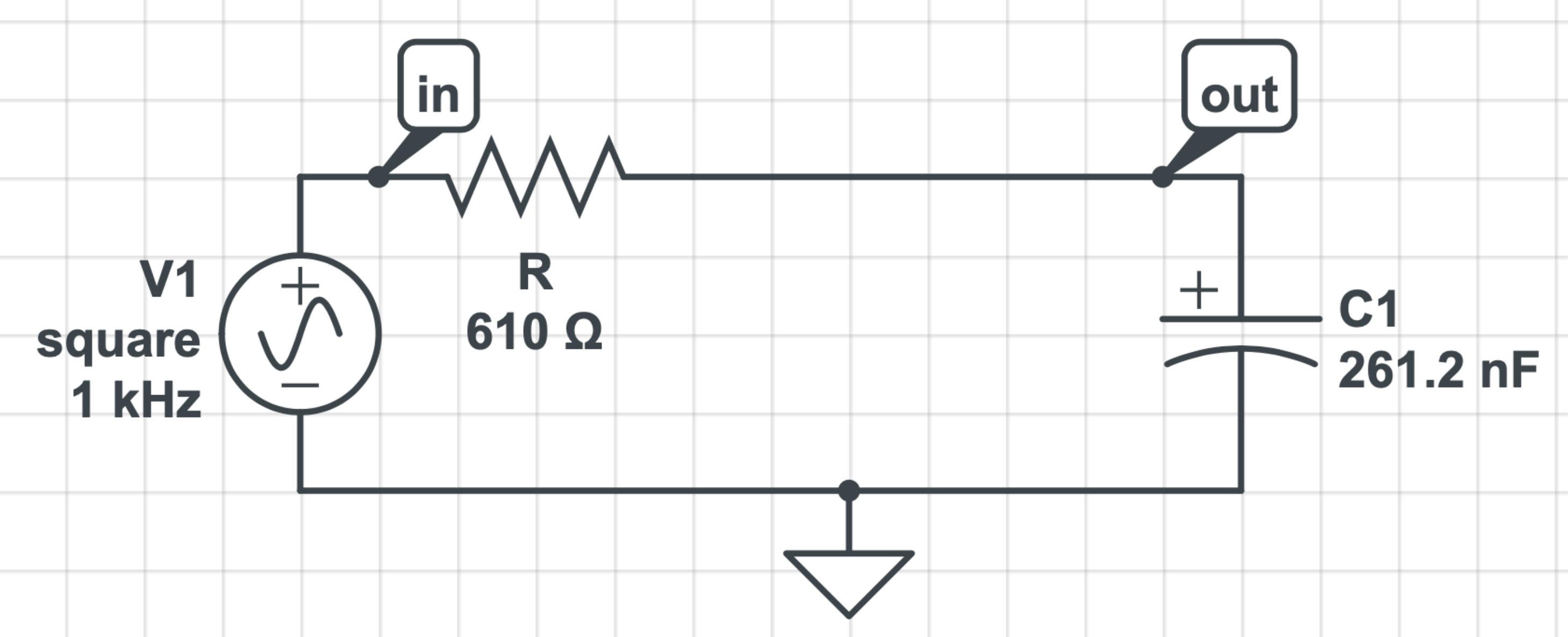}
    \centering
    \caption{Low-pass filter architecture (after human intervention on the GA schematic) at the end 
of the second step.}
\label{fig:CNMALPF}
\end{figure}

We perform fine-tuning through an NN with a hidden layer consisting of 100 neurons 
that is converted into an MILP. The NN inputs are component values (resistor and capacitor) 
that are either derived from a feasible solution to the MILP or else a random sample if a feasible 
solution does not exist. The continuous search space is $[400, 800] \Omega$ for \textit{resistor}
and $[0.01, 1]\rm \mu F$ for \textit{capacitance}. Since this is a simple design, 
through prior knowledge we know that a low-pass filter can be synthesized within these ranges of 
component values to meet the specifications. The targeted outputs of the NN are gain, bandwidth, and 
response of the filter at 200, 500, and 2000 Hz. The responses at these frequencies sufficiently 
capture the behavior of the filter below and above the desired bandwidth. The output constraints 
are as follows: gain (from input to output) in the range $[-0.92, 0.83]$ dB and bandwidth in the range $[990, 1010]$
Hz. We generate 10 random input samples to initialize the NN for training to 
minimize the mean-squared error.  The solutions suggested by MILP meet the requirements after 20 
more simulations. We show these simulations (10 for initialization of the NN and 20 during the MILP step) 
in Fig.~\ref{fig:CNMASimulation}: initialization points in the orange part and MILP ones in the blue 
part.  The outcomes during initialization can be seen to be random as they are far from meeting the 
requirements. However, the points suggested by the MILP have a response closer to the requirements 
until finally the suggested point meets the specification.  The circuit with the values of the 
components on termination of the second step is shown in Fig.~\ref{fig:CNMALPF}. 

\section{Synthesis Methodology}
\label{sect:methodology}
In this section, we describe DISPATCH in detail.  The first step 
explores a large search space using GA based on gradient-free search. The outcome of the first step 
is a \emph{coarse} design that is fine-tuned through gradient-based, sample-efficient search to 
obtain better component values.

\subsection{Step 1: Coarse design}
\label{sec: coarseDesign}
\begin{figure}[h]
    \includegraphics[scale=.46]{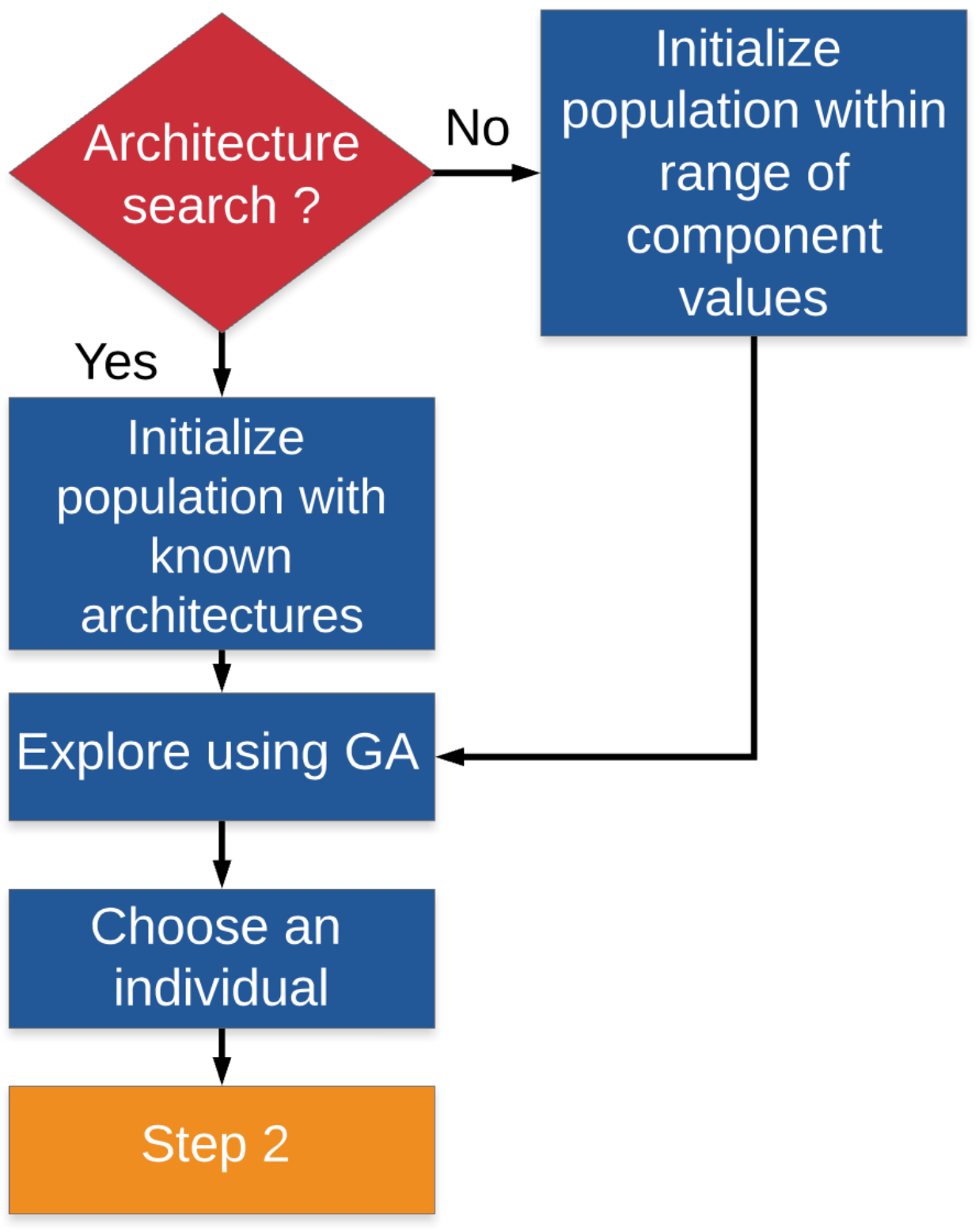}
    \centering
    \caption{Evolution using GA.}
\label{fig:Step 1 GA}
\end{figure}

We show the flow involved in the synthesis of the coarse design in Fig.~\ref{fig:Step 1 GA}. We 
evolve individuals across generations using NSGA-\RNum{2} \cite{996017} that yields non-dominated 
solutions of the CPS design that is formulated as an MOO problem.  In architecture search, we initialize some individuals in the first 
GA generation with seed designs from the literature to exploit prior knowledge. When performing only 
component selection, we initialize the individuals in the first generation with component values to 
cover the search space.  This enables comparisons with other designs from the literature that do not 
use prior knowledge.

During \emph{{architecture search}}, a \emph{gene} has three constituents: type of component, 
nodes connecting the component, and the component value, as shown in 
Fig.~\ref{fig:LowPassGene}. In the case of \emph{{component selection}}, the gene encodes the value of the component in the \emph{chromosome}
that represents the 
circuit. We evaluate each circuit represented by a chromosome using \emph{HSPICE}. 

Algorithm \ref{alg:GAArchAlgorithm} describes how architecture search is performed using GA. We 
generate \textit{P} individuals in a generation. Some of these are from seed designs (\textit{S}) 
whereas others are randomly generated.  We generate random individuals by selecting a random 
component, choosing the number of nodes (e.g., a resistor requires two nodes whereas a MOSFET 
requires three nodes) connected to that component, and the component value. We select the component 
values for random individuals from quasi-random \textit{Sobol} samples in the range (\textit{R}) of 
each component. Sobol samples are distributed uniformly over a unit hypercube 
\cite{burhenne2011sampling}.  Then we scale the samples to lie within the specified range for each 
component. We postprocess individuals to ensure that some components are connected to the fixed 
terminals of the architecture, e.g., \emph {ground, supply voltage, input/output terminals in
the case of a circuit}, to ensure a valid design. We simulate all individuals in a generation to 
compute the objective functions. This is followed by ranking using the \textit{NSGA-\RNum{2}} 
algorithm \cite{996017}. We use tournament selection for selecting some individuals in a generation 
to undergo reproduction through \emph{crossover} and \emph{mutation} to produce \textit{P} children.
We use \textit{NSGA-\RNum{2}} again to select \textit{P} individuals out of the \textit{2P} 
individuals for the next generation. This process continues until one of the stopping criteria
(\textit{stop}) is met. These criteria are based on individuals not improving over a fixed number 
of generations, exhausting the simulation budget or on attaining the required performance. Finally, 
we select one individual from the final generation based on a CPS performance metric. The metric 
(lower the better) in our case is one of the multiple CPS objectives. We select an individual from the 
final generation since ranking using \textit{NSGA-\RNum{2}} ensures that we never lose an individual from the Pareto front within a generation. We select only one individual since GA only synthesizes a coarse CPS design although it is also possible to select another individual to undergo 
fine-tuning in the next step. 

\begin{algorithm}[h]
    \caption{Step 1: Architecture synthesis using GA}
    \label{alg:GAArchAlgorithm}
    \begin{algorithmic}[l]
        \REQUIRE \textit{S}: Seed design(s); \textit{N}: $\#$generations; \textit{P}: population size; \textit{C}: Components; \textit{R}: Component range; \textit{nodes}: Nodes; \textit{max\_comp}: Max $\#$ components; \textit{stop}: Stopping criteria
        \STATE - Generate \textit{Sobol} samples for \textit{C} within \textit{R}
        \STATE - Initialize individuals with \textit{S} and others randomly
        \WHILE{not \textit{stop}} 
            \STATE - Compute \textit{objectives} for all \textit{P}
            \STATE - Rank \textit{P} using \emph{NSGA-\RNum{2}}
            \STATE - Use tournament selection to create mating population of size \textit{P}
            \STATE - Use reproduction based on \emph{crossover} and \emph{mutation} to create 
\textit{P} children
            \STATE - Select \textit{P} from \textit{2P} members using \emph{NSGA-\RNum{2}}
        \ENDWHILE
        \ENSURE Best individual from the final generation using a metric.
    \end{algorithmic}
\end{algorithm}

Component selection for a fixed architecture using GA is done using a setup similar to Algorithm \ref{alg:GAArchAlgorithm}. 
Instead of initializing some individuals in the first generation by seed designs, we initialize all 
the individuals using Sobol samples. The rest of the procedure remains the same.

\subsection{Step 2: Fine-tuning}
\label{CNMA}
\begin{figure}[h]
    \includegraphics[scale=0.23]{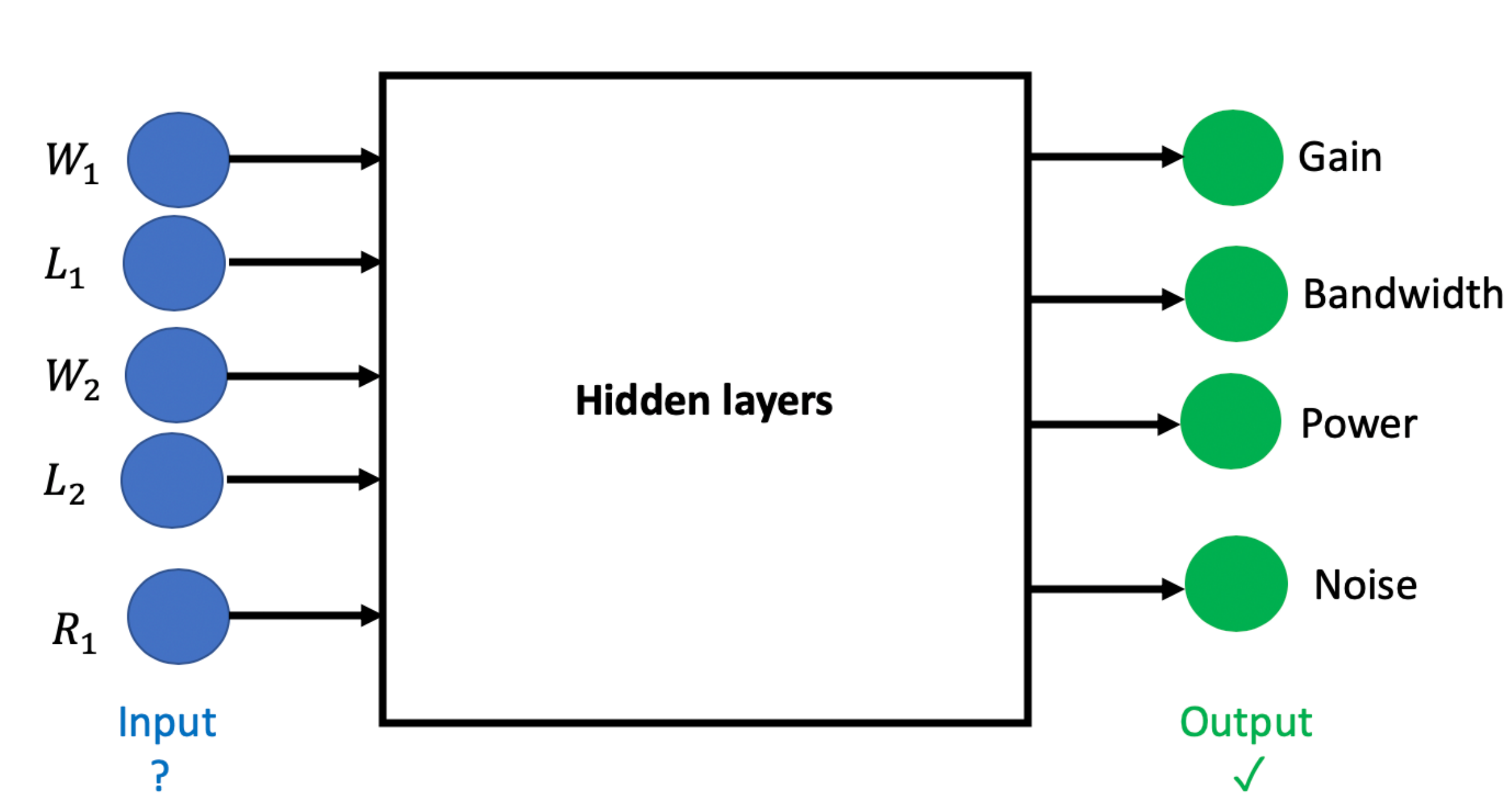}
    \centering
    \caption{Overview of gradient-based search through an example.} 
\label{fig:CNMAHighLevelWhy}
\end{figure}

At the end of Step 1, we have a coarse CPS design. When using Step 1 for architecture search, 
we use human intervention to refine the synthesized design if needed. We fine-tune this design
through a modified version of CNMA \cite{CNMA} by searching in a continuous space. In contrast to 
CNMA where the selection is stopped when the simulation budget is exhausted, we adaptively replace the 
design from Step 1 if a successful design is not found within a fixed number of simulations. We model the 
response of the system to the inputs using an NN.   The NN converted to MILP acts as an 
\emph{acquisition function} and determines the sample point for CPS simulation. We train the NN to 
minimize the mean-squared error of the CPS response to the inputs.
Fig.~\ref{fig:CNMAHighLevelWhy} shows the high-level overview of this step through an example. The 
designer specifies the system requirements shown on the right in green. The feasible solution of the MILP 
determines the potential component values shown on the left in blue that achieve the desired response.

\begin{figure}[!ht]
    \includegraphics[scale=.323, angle =270]{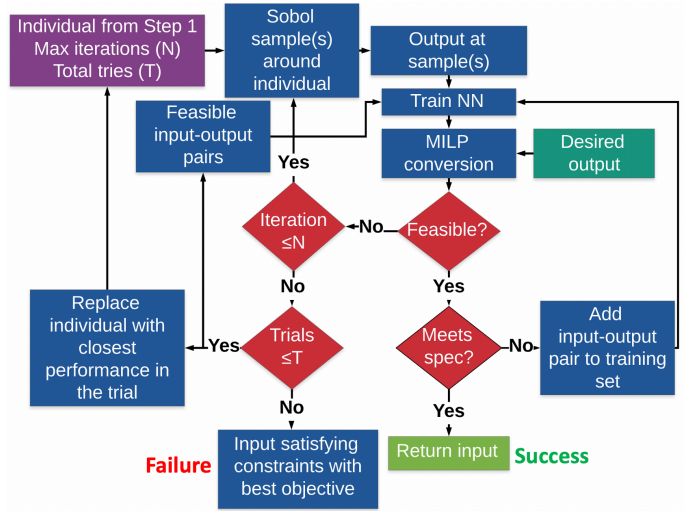}
    \centering
    \caption{Flowchart that illustrates the fine-tuning procedure in Step 2 of DISPATCH.}
\label{fig:CNMAFlowchart}
\end{figure}

Fig.~\ref{fig:CNMAFlowchart} shows the procedure to fine-tune the design that Step 1 yields. 
In the flowchart, \textit{N} denotes the number of simulations in a particular trial out of a total 
of \textit{T} trials. We effectively repeat the fine-tuning step \textit{T} times, allowing a maximum 
of \textit{N} simulations in a trial.  We obtain \textit{Sobol} samples from the range of each 
component in the design from Step 1, followed by simulation of these points to determine the output. 
The samples are generated around the nominal values of each components (e.g., $\pm 70\%$ of 10 
$\Omega$ for resistor). The samples are clipped to ensure that the values of the components are within the permissible range for each type of component.  We train
an NN with these input-output pairs that are range-normalized to $[0,1]$. We convert the NN into
an MILP using Eq.~(\ref{eq:MILPConversion}) to see if there is a feasible solution. The range of 
available components determines the constraints on the input whereas system requirements determine 
the output constraints.  We use Gurobi \cite{gurobi} to find a feasible solution.  If such a solution 
exists, it indicates that the desired output is reachable by the NN from this input. This corresponds 
to a feasible solution.  We simulate the system with the suggested input. The simulation terminates 
if the system requirements are met. Otherwise, we use the input-output pair to train the NN further.
The procedure continues until a maximum number of permissible simulations (\textit{N}) is exhausted. 
After finishing a trial, the input corresponding to the least absolute sum in terms of fractional 
deviation of the output from the requirement replaces the design from Step 1. Fractional deviation 
is computed as follows:
\begin{equation}
\label{eq: MinDeviationCNMA}
    \begin{aligned}
    \sum_{i}\frac{\left|{\rm obs}_i\ -\ {\rm spec}_i\ \right|}{{\rm spec}_i}\ \mathds{1}_{{\rm \{obs_{i} \lessgtr spec_{i}\}}},
    \end{aligned}
\end{equation}

\noindent
where ${\rm spec}_i$ is the specification for the $i^{th}$ objective or constraint, ${\rm
obs}_{i}$ is the observed response of the CPS for the $i^{th}$ objective or constraint. We use the inputs that yield the minimum fractional deviation over the sum of all the objectives in a trial.
$\mathds{1}_{{\rm \{obs_{i} \lessgtr spec_{i}\}}}$ is an indicator function that takes the value 1 in the case of 
violation of ${\rm spec}_i$ determined by ${\rm obs}_i$ (can be greater or less than) and 0 otherwise.  We retain the feasible solutions obtained from the 
MILP formulation to further train the NN in subsequent trials. After the last trial, we return the 
input that corresponds to the output that satisfies all the hard constraints (like maximum power 
consumption for a valid design) and comes closest to satisfying the target objective of the system. This  indicates a failure as the system does not meet the requirements exactly.

\section{Experimental Results}
\label{sect:results}
In this section, we evaluate how DISPATCH performs architecture search and component selection. 
We first show the architecture search and component selection  results for a \textit{two-stage transimpedence amplifier} 
and then the component search results for both \textit{two-stage} and \textit{three-stage} 
transimpedence amplifiers. We compare our results with those in \cite{wang2018learning} that are 
synthesized by humans, RL, and Bayesian optimization. The technology files specifying the device 
physics for simulation are from \cite{NIPS2019_8519} and are the same as in
\cite{wang2018learning}. We implement DISPATCH using Keras \cite{chollet2015keras},
scikit-learn \cite{scikit-learn}, Gurobi \cite{gurobi}, and PyGMO \cite{pygm}. The simulations
are performed on an Intel Xeon processor with 128 GB of DRAM. 

\subsection{Architecture search and component selection}
\label{subsec:ArchExample}
\begin{figure}[h]
    \includegraphics[scale=0.3]{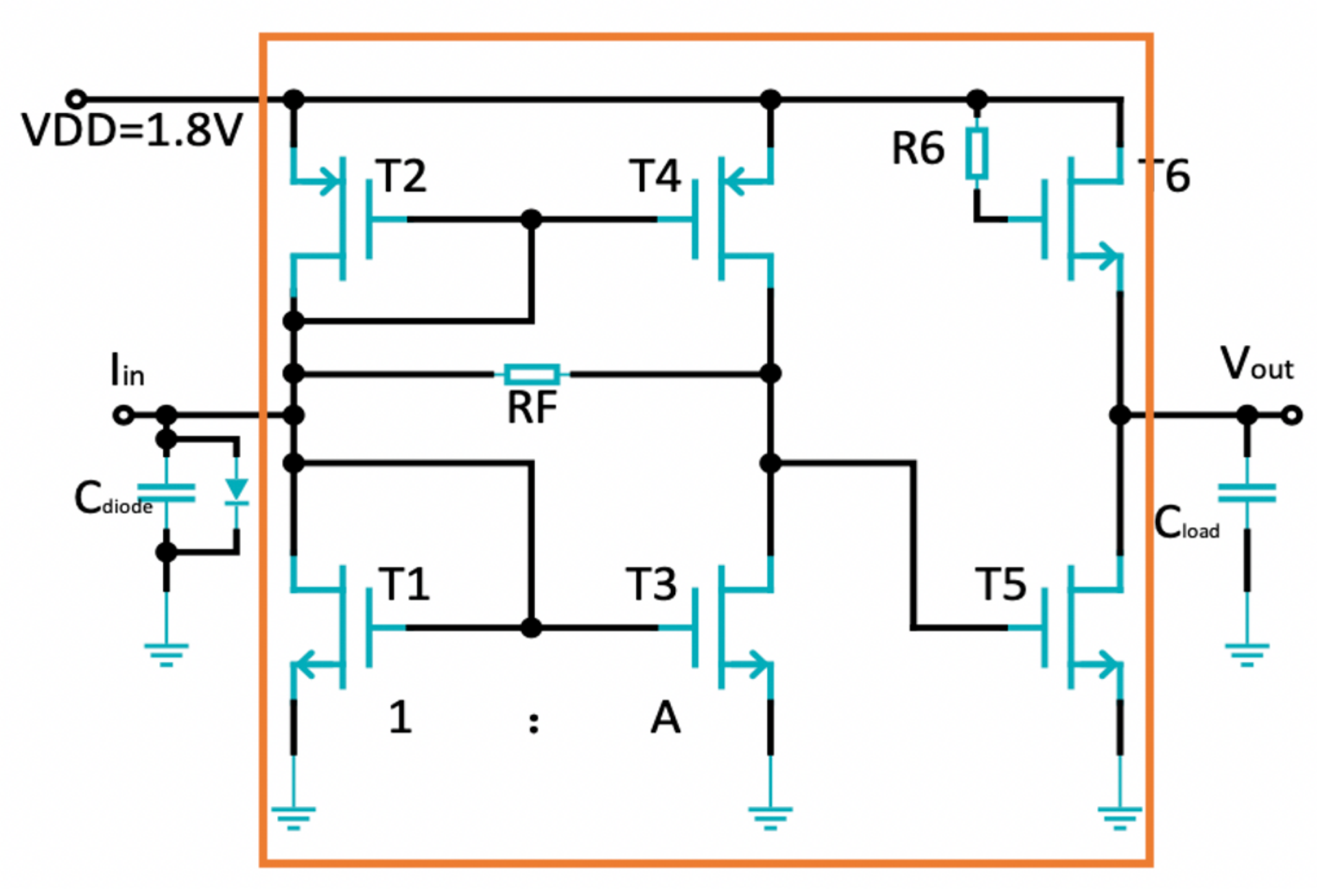}
    \centering
        \caption{Two-stage transimpedence amplifier from \cite{wang2018learning}. Component values are selected for devices inside the orange box and the outside is fixed.}
\label{fig:twoStageSeed}
\end{figure}

We use the standard design of a two-stage transimpedance amplifier presented in 
\cite{wang2018learning} as the seed design for our methodology in order to take advantage of prior 
human knowledge and make improvements in all the metrics.  Fig.~\ref{fig:twoStageSeed} shows this seed design. The objective is to maximize the bandwidth of the amplifier while minimizing the sum of gate areas of all the MOSFETs in the circuit and satisfying 
hard constraints on \textit{noise, gain, peaking}, and \textit{power}. The search space comprises 
three components: two types of MOSFETs (PMOS transistor, NMOS transistor) and resistor. In the 
human-designed circuit, the 
MOSFETs are of minimum length ($0.18\ \mu \rm m$), based on the technology used, whereas the width is 
variable. Our search also uses minimum-length MOSFETs and only selects their width.  We
discretize the search space during architecture search (Step 1) by generating 100 Sobol samples for 
resistors in the $[50, 5\rm k]\ \Omega$ range and width in the $[0.18, 80]\ \mu \rm m$ range. 
In the
human-designed circuit, the resistor values are $420~\Omega$ and $3~\rm k\Omega$, and
the width is in the $[0.9, 51]\ \mu \rm m$ range. During evolution, we allow the circuit to have a 
maximum of 11 nodes and a total of 10 components, whereas the seed design has six nodes and eight 
components, to enable search for novel designs. Since we need 10 components for a chromosome, 
we initialize two additional genes as resistors with a small resistance of 2.2 $\rm \mu \Omega$ whose terminals are shorted and not connected to any of the nodes in the seed design. This ensures the response of the seed design remains unaffected.

\begin{figure}[!ht]
    \includegraphics[scale=0.3]{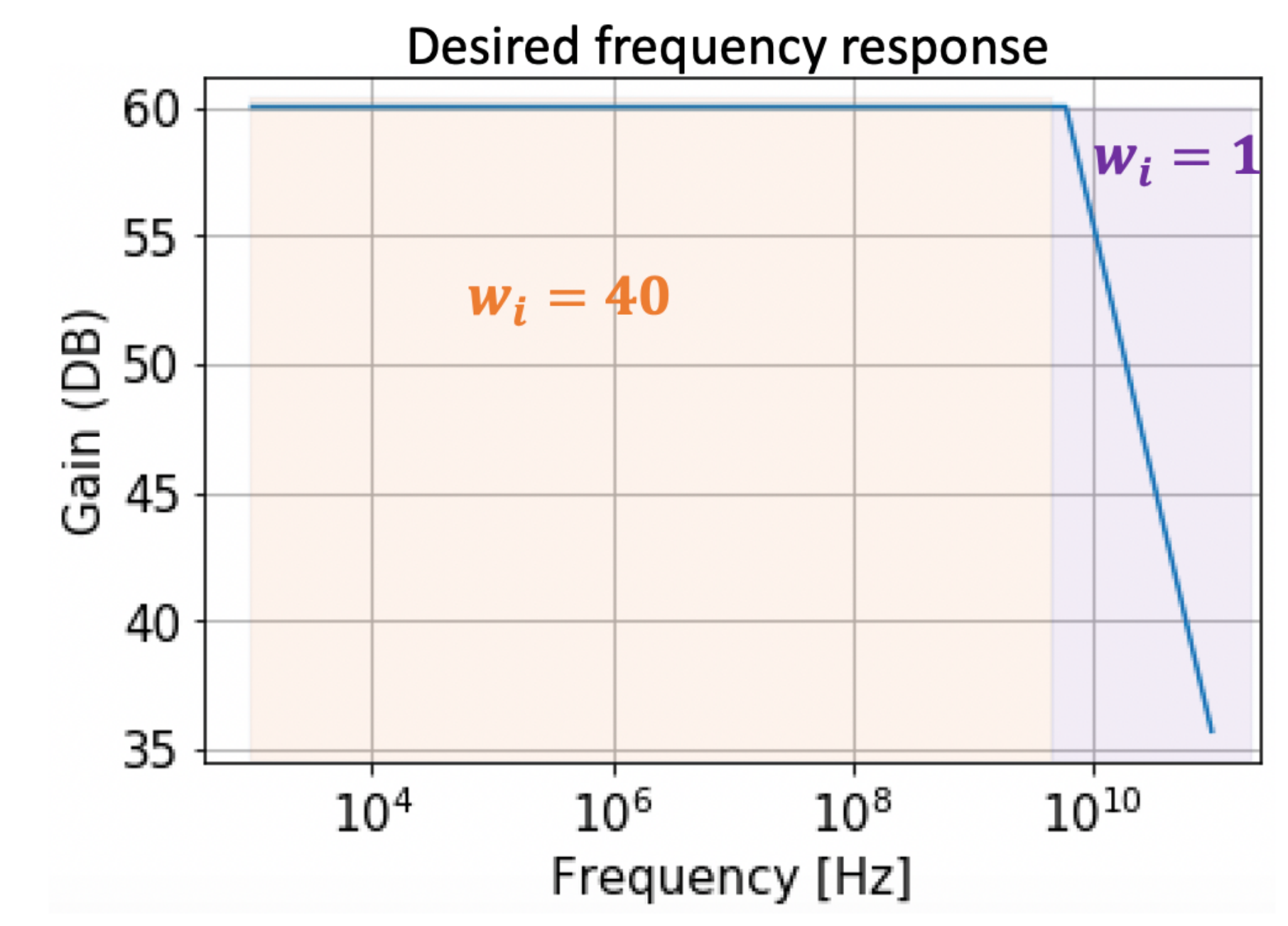}
    \centering
    \caption{Desired frequency response of the circuit. The orange-shaded region represents the 
passband and the purple-shaded region the stopband. $w_{i}$ shows the weights in each region. }
\label{fig:desiredResponse}
\end{figure}

We use GA to evolve a generation of 100 individuals and set the maximum number of generations to 200. 
We choose these numbers to achieve at least about $50\%$ sample efficiency (i.e., $50\%$ fewer 
simulations) compared to the design methodology in \cite{wang2018learning}. We simultaneously search for the 
architecture and the component values whereas the method in \cite{wang2018learning} only searches for 
component values. Hence, our method searches through a much larger design space, albeit with the seed design. We formulate an MOO problem with three objectives: bandwidth, noise, and power.  The objectives are described next.
\begin{enumerate}
    \item \textit{Bandwidth}: This objective corresponds to the weighted sum of the absolute 
difference between the desired and observed responses, with a passband weight of 40 and a
stopband weight of 1, as shown in Fig.~\ref{fig:desiredResponse} with weights ($w_i$) in each region 
in text. 
A reward/penalty is applied to this objective as follows.
    \begin{itemize}
        \item Assessing the level of reward/penalty is based on the operating region of the MOSFET: 
reward for MOSFET operating in the saturation region, else a penalty, as follows:
        \begin{itemize}
            \item \textit{Saturation region}: Reward of 1 divided by the number of MOSFETs. 
            \item \textit{Linear region}: Penalty of 2 divided by the number of MOSFETs.
            \item \textit{Cutoff region}: Penalty of 3 divided by the number of MOSFETs.
        \end{itemize}
        The penalty encourages MOSFET operation in the saturation region. 
        \item A penalty of 15 is assessed based on the fractional deviation in gain below 
58.1 dB (since 58.1 dB is the gain achieved by RL based synthesis in 
\cite{wang2018learning}).
        \item A penalty of 15 is assessed based on the fractional deviation in peaking above 0.963 
(achieved by RL in \cite{wang2018learning}).
        \item A penalty of 15 is assessed based on the fractional deviation in bandwidth below 
5.81 GHz (this is slightly higher than the bandwidth achieved by RL in 
\cite{wang2018learning}: 5.78 GHz).
    \end{itemize}
    We scale the bandwidth objective by dividing it by the objective of the seed design.
    \item \textit{Noise}: This objective corresponds to the ratio of measured noise and the noise 
achieved by the RL-based design in \cite{wang2018learning} (19.2 $\rm pA/\sqrt{\rm Hz}$). An additional 
penalty of 15 is assessed based on the fractional deviation in noise above this value.
    \item \textit{Power}: This objective corresponds to the ratio of measured power and the power 
achieved by the RL-based design in \cite{wang2018learning} (3.18 $\rm mW$). No penalty is 
assessed in this case due to the large room for optimization available for power consumption.
\end{enumerate}

\noindent
The objective for noise is given by,
\begin{equation}
    \label{eq: NoiseObj}
    \ \frac{N_m}{N_{ref}}+\alpha\frac{N_m\ -\ N_{ref}\ }{N_{ref}}\mathds{1}_{\{N_m>N_{ref}\}},
\end{equation}
\noindent
where $N_{ref}$ is the noise of RL-designed circuit, $N_{m}$ is the measured noise, $\alpha$ is
the penalty factor (15 in our case) and $\mathds{1}_{\{N_m>N_{ref}\}}$ is the indicator
function that takes a value of 1 if $N_m>N_{ref}$ and 0 otherwise. The second term is a penalty 
applied only when the measured noise ($N_{m}$) is worse than $N_{ref}$. The total penalty scales 
linearly with deviation from $N_{ref}$.  Other objectives are defined analogously.

We club together all the sub-objectives related to bandwidth (gain, desired bandwidth, peaking) into one to minimize the number of 
objectives that need to be tackled while capturing the entire frequency response. The penalty 
terms encourage the target response to be better than the response of the designs in \cite{wang2018learning}. We 
do not tune the penalty to the circuit since the aim of Step 1 is to synthesize a coarse design. In 
case the simulation is unsuccessful for a circuit that does not meet the objectives, we set
the objective values to a very large number to indicate this fact.

We use a tournament size of 10, mutation rate of 0.1, and crossover probability of 0.9 that are 
typical values for GA. The evolution stops under any of the following circumstances:
\begin{itemize}
    \item The scaled bandwidth objective falls below 0.9 since the aim is to improve it by about 
$10\%$ relative to that of the seed design. 
    \item The number of generations exceeds 100 and the bandwidth objective stays the same for more 
than 100 generations, indicating saturation in GA performance.
    \item The maximum number of generations (200) is reached.
\end{itemize}

\begin{figure*}[htbp]
\centering
\subfloat[]{\includegraphics[height=1.3in, width=2.in]{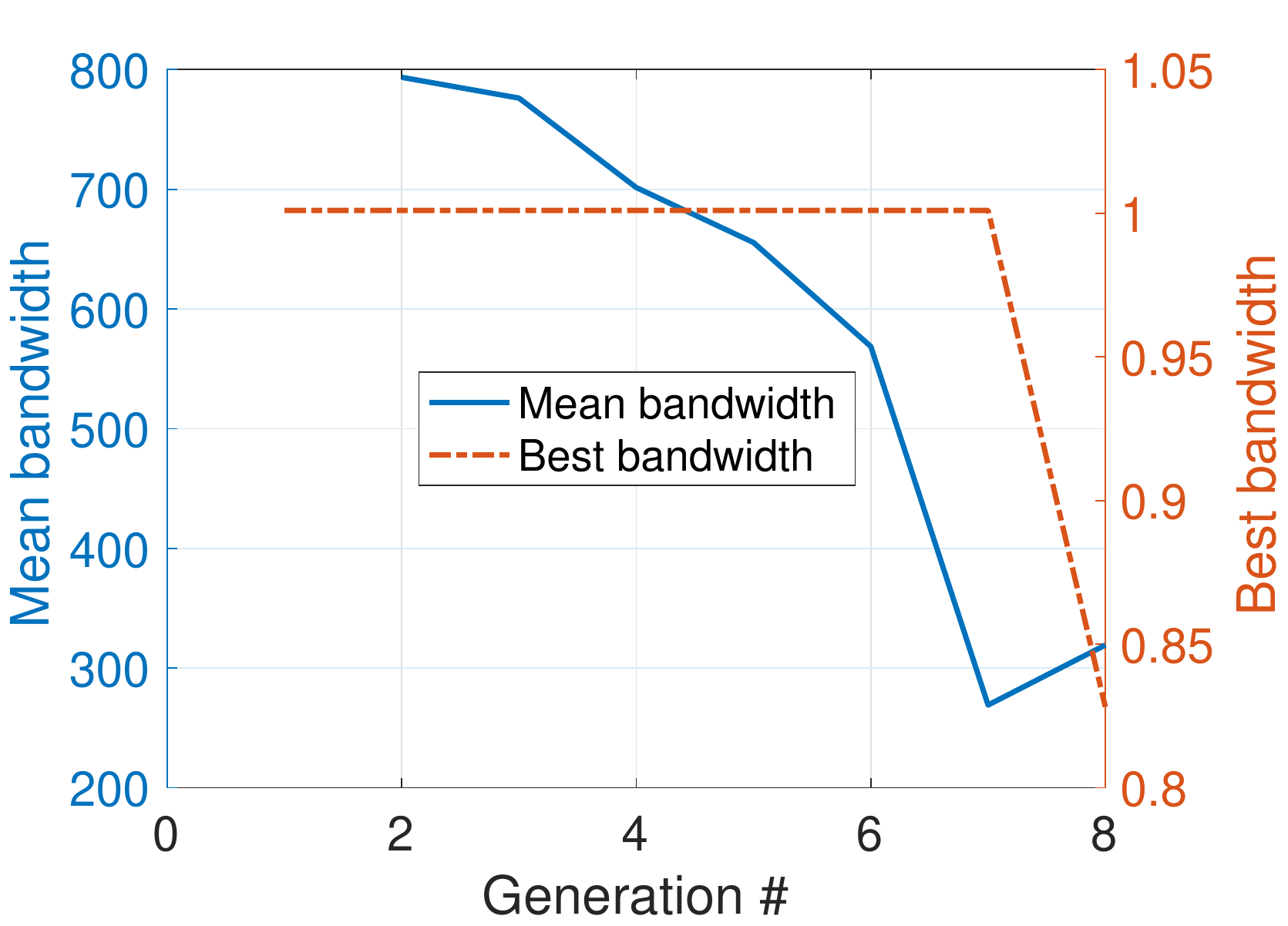}}
\subfloat[]{\includegraphics[height=1.3in, width=2.in]{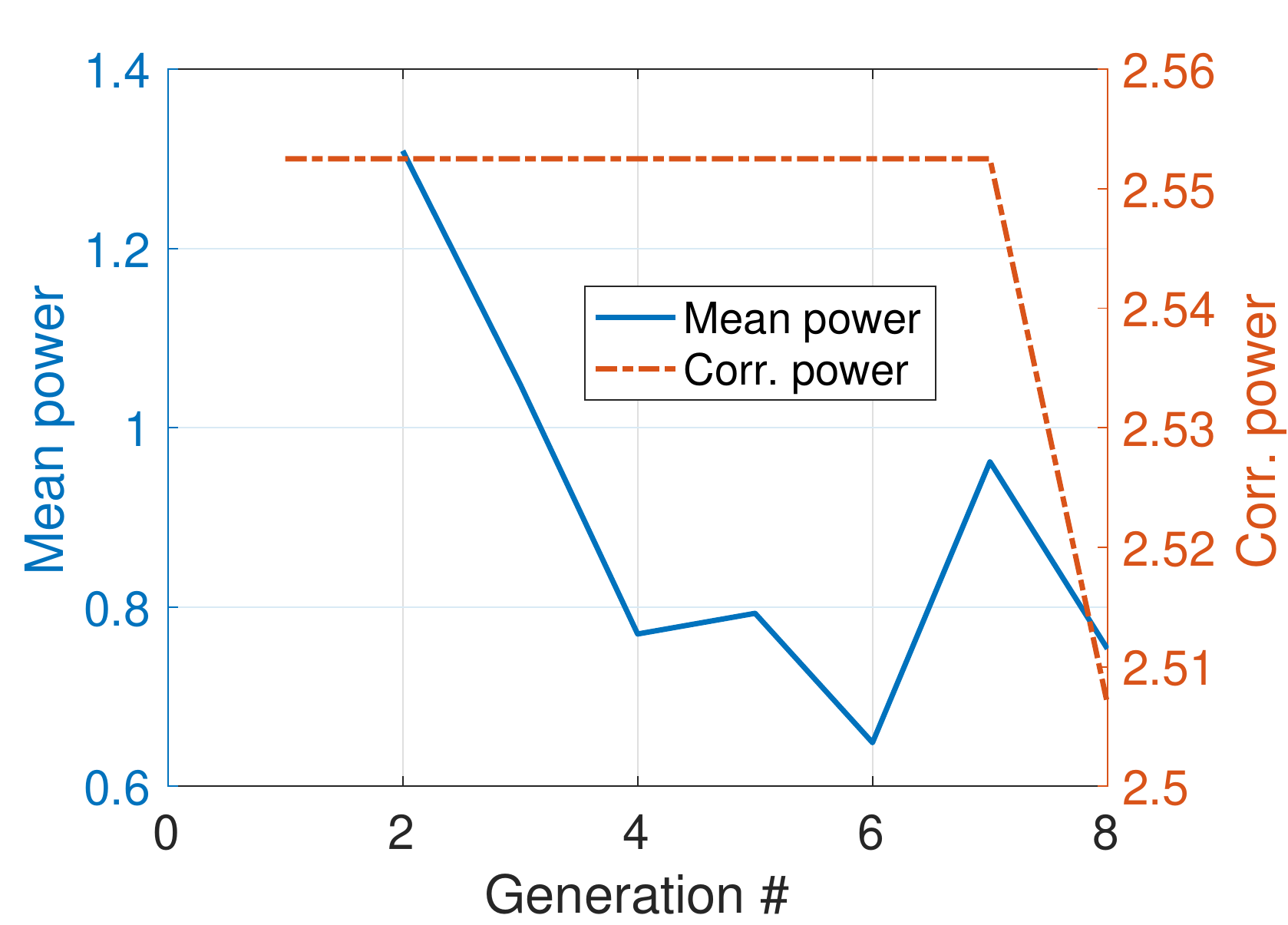}}
\subfloat[]{\includegraphics[height=1.3in, width=2.in]{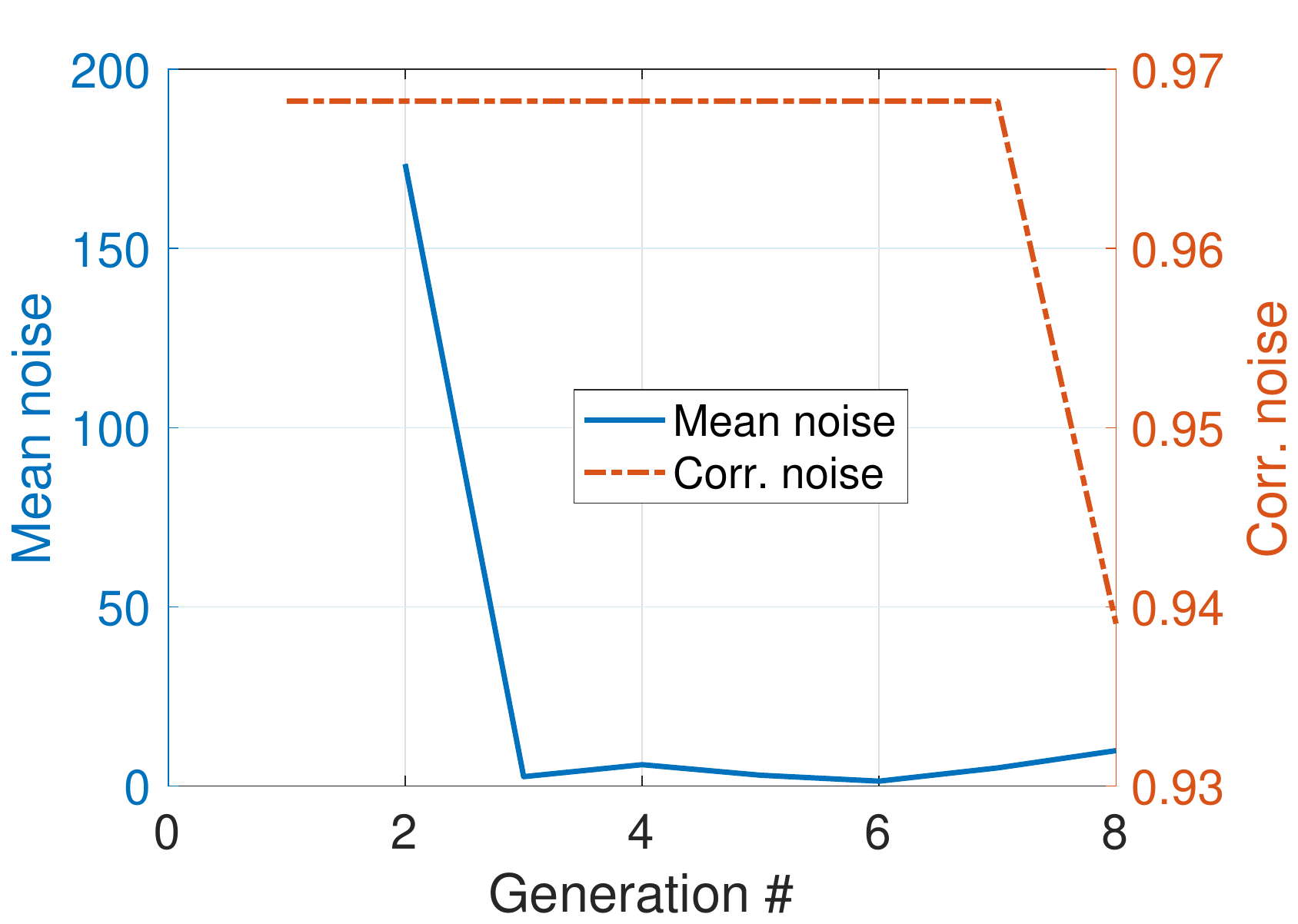}}
\caption{Scaled objectives across generation for (a) bandwidth, (b) power, and (c) noise from 
the second generation onwards for architecture synthesis of the two-stage 
transimpedence amplifier. The solid blue line shows the mean objectives averaged across 
all individuals in a generation and the dotted-orange line shows the individual with the best 
objective for (a) bandwidth and the corresponding (Corr.) objectives for (b) power, and (c) noise.}
\label{fig:GAAnalysisArch}
\end{figure*}

Fig.~\ref{fig:GAAnalysisArch} shows the evolution of objective values across GA generations. 
It depicts the mean values of all the objectives i.e., (a) bandwidth, (b) power, and (c) noise, 
across all individuals in a generation after the second generation, shown in blue. As the $y$-axis 
shows scaled objectives, it has no units. We plot from second generation onwards due to the high 
objective values in the first generation.  There is a trade-off among the 
three objectives. The dotted-orange curve shows the individual with the best objective for bandwidth 
in each generation and the corresponding objectives for power and noise in
Fig.~\ref{fig:GAAnalysisArch}(b) and Fig.~\ref{fig:GAAnalysisArch}(c), respectively. In the
first seven generations, the seed design with a scaled bandwidth objective of 1 is the lowest among 
all the designs generated by GA. In the eighth generation, a design is evolved with a lower objective value for bandwidth compared to the seed design. The simulation stops at this point as the bandwidth objective is below 0.9, thereby meeting the stopping criteria.

\begin{figure}[ht]
\includegraphics[scale=0.2, angle = 0]{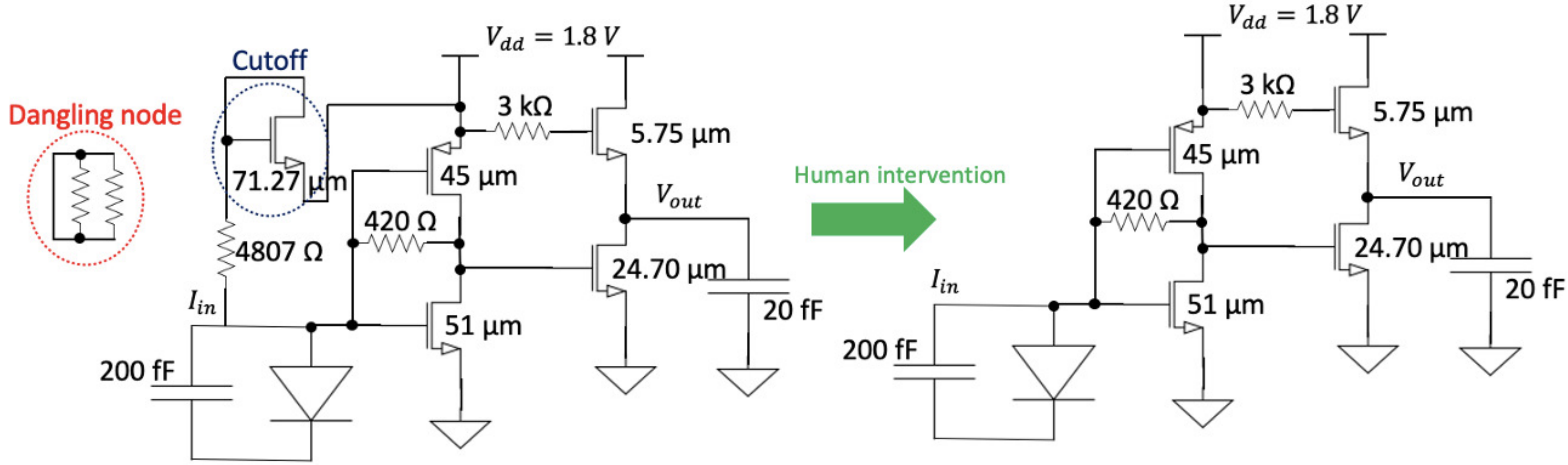}
\centering
\caption{Circuit synthesized by GA followed by human intervention. All MOSFETs are of minimum length (0.18 $\mu m$).}
\label{fig:GACircuit}
\end{figure}

We choose the circuit with the best objective for bandwidth as the coarse design in Step 1 since this 
circuit addresses several sub-objectives.  We fine-tune the coarse design to meet the specifications 
in the next step. We remove dangling nodes through human intervention as they are redundant. We also 
remove a MOSFET operating in the cutoff region as it does not contribute to gain. 
Fig.~\ref{fig:GACircuit} shows the circuit 
before and after human intervention. Table \ref{tab:GAvsHumanTable} shows the values of objectives and constraints 
before and after human intervention. GA requires less than an hour for synthesis. 
After human intervention, the circuit does not meet the hard constraint for peaking. 
This can be remedied in Step 2 through fine-tuning.  The seed design has six MOSFETs. This one uses only four MOSFETs.

\begin{table*}[h]
\centering
\caption{Comparison of the GA-synthesized circuit with the human-designed circuit for architecture 
search for the two-stage transimpedence amplifier. Hard constraint violations are shown in a circle.}
\label{tab:GAvsHumanTable}
\resizebox{\textwidth}{!}{\begin{tabular}{|c|c|l|c|c|c|c|c|c|}
\hline
 & \#Samples & Time & Noise ($pA/\sqrt(Hz)$) & Gain ($dB\  \Omega$) & Peaking (dB) & Power (mW) & Gate area (${\mu m}^2$)  & Bandwidth ($GHz$) \\ \hline
Spec.         & -       & -                      & $\leq 19.3$ & $\geq 57.6$ & $\leq 1$ & $\leq 18$ & -       & maximize    \\ \hline
Human Design \cite{wang2018learning} & 1,289,618 & months                 & 18.6         & 57.7         & 0.927    & 8.11
& 23.11 & 5.95 \\ \hline
GA           & 1,600    & 0.15 hr                & 18.0         & 58.0         & 0.913    & 7.97       & 35.59   & 5.93 \\ \hline
GA+Human     & -    & \multicolumn{1}{c|}{-} & 17.9         & 58.0          & \enumber{1.042}     & 7.97       & 22.76   & 5.96 \\ \hline
\end{tabular}}
\end{table*}

The next step is selecting the component values to meet the specifications. In addition, our
goal is to have the synthesized circuit dominate all the designs in \cite{wang2018learning}.
In Step 2, we specify the range of each component in the circuit in the first iteration to $70\%$ below and
above the value of the component selected in Step 1, provided these values are within 
the range of permissible values, else we clip the range to permissible values.
The maximum number of simulations we allow in the first trial, besides NN initialization, is 200. 
We increase it to 500 for each subsequent trial. The intuition behind the choice is that the NN 
learns a better system representation over time and is, hence, permitted to make more guesses in
subsequent trials. 
We derive 100 Sobol samples based on the circuit synthesized in Step 1 to train the NN in the first trial. For subsequent trials, we only use one Sobol sample based on the design selected in the previous trial. We use the remaining feasible points from previous iterations to train the NN before formulating the MILP in Step 2. The NN has three hidden layers with 40, 20, and 8
neurons, respectively. We generate training data over time, thereby making it robust to the choice 
of the NN as long as the NN can model sufficient complexity. 
We use \textit{MLP Regressor} from the Scikit-learn \cite{scikit-learn} package along with 
the Adam optimizer \cite{duchi2011adaptive}, an initial learning rate of 0.0001, 
\textit{adaptive} learning, and a \textit{maximum iteration count} of 100000 to train the NN. Other parameters are set to their default values.
We run Step 2 for a maximum of 144 hours or 20 trials whichever occurs first. Unless specified, the same setup is used for all the experiments.

Table \ref{tab:GAArchComparisonPaper} \footnote{Gate area is shown only for designs for which
this information is available. There was a problem in area calculation in \cite{wang2018learning} 
that was confirmed after contacting the authors.} shows a comparison of designs synthesized using our 
methodology with designs obtained by humans, DDPG, and Bayesian optimization 
\cite{wang2018learning}. The last row depicts a design that satisfies all the hard constraints while 
maximizing bandwidth.
We save simulation results at the end of every trial. The time column 
includes the time required for Step 1 and
the time elapsed until the last saved trial. The number of samples simulated is cumulative 
over the two steps (GA and fine-tuning). For the designs from \cite{wang2018learning}, the
number of samples only accounts for component selection. DISPATCH synthesizes designs that are better at meeting the objectives across the board relative 
to the weighted sum approach taken in \cite{wang2018learning}.
We also obtain a valid design with the highest bandwidth compared to all the designs in 
\cite{wang2018learning}. This bandwidth corresponds to the design that meets all the hard constraints 
and has the highest bandwidth among all the simulations performed.  There is also a significant 
reduction in synthesis time: ours in CPU hours and for the designs from \cite{wang2018learning} in 
GPU hours. We set the required area to be less than $0.85\times$ of the human design when dominating 
the human design, less than $0.75\times$ of the human design when dominating the DDPG design, less 
than $0.85\times$ of the human design when dominating the Bayesian optimization based design, and 
less than $0.9\times$ of the human design when maximizing bandwidth. The correct area is not available 
for all the designs in \cite{wang2018learning}. Hence, we use the human-designed circuit as a 
reference for area computation.

\begin{table*}[h]
\centering
\caption{Comparison of circuits synthesized with DISPATCH and designs in 
\cite{wang2018learning}.  Hard constraint violations are shown in a circle. }
\label{tab:GAArchComparisonPaper}
\resizebox{\textwidth}{!}{\begin{tabular}{|c|c|l|c|c|c|c|c|c|}
\hline
 &
  \#Samples &
  Time &
  Noise ($\rm pA/\sqrt{\rm Hz}$) &
  Gain ($\rm dB\  \Omega$) &
  Peaking ($\rm dB$) &
  Power ($\rm mW$) &
  Gate area (${\rm \mu m}^2$) &
  Bandwidth ($\rm GHz$) \\ \hline
Spec.          & -       & -        & $\leq 19.3$ & $\geq 57.6$ & $\leq 1$ & $\leq 18$ & -       & maximize    \\ \hline
Human Design \cite{wang2018learning} & 1,289,618 & months & 18.6         & 57.7         & 0.927    & 8.11      & 23.11 & 5.95 \\ \hline
{Human Design (DISPATCH)} &
  {1,911} &
  {1.2 hrs} &
  {18.4} &
  {57.7} &
  {0.878} &
  {5.77} &
  {17.91} &
  {6.00} \\ \hline
DDPG  \cite{wang2018learning}        & 50,000   & 30 hrs & 19.2         & 58.1         & 0.963     & 3.18       & -       & 5.78 \\ \hline
{DDPG (DISPATCH)} &
 {2,291} &
  {4.7 hrs} &
 {19.2} &
  {58.1} &
 {0.953} &
  {3.16} &
  {11.81} &
  {5.80} \\ \hline
Bayesian Opt.\cite{wang2018learning} & 880     & 30 hrs & \enumber{19.6}         & 58.6         & 0.629     & 4.24       & -       & 5.16 \\ \hline
{Bayesian Opt. (DISPATCH)} &
  {1,751} &
  {0.5 hr} &
  {17.6} &
  {58.8} &
  {0.260} &
  {3.84} &
  {15.25} &
  {5.23} \\ \hline
{Max. Bandwidth (DISPATCH)} &
  {4,176} &
  {106.3 hrs} &
  {19.2} &
  {57.6} &
  {0.984} &
  {5.85} &
  {20.13} &
 {6.18} \\ \hline
\end{tabular}}
\end{table*}

\begin{figure*}[!ht]
    \includegraphics[scale=0.33]{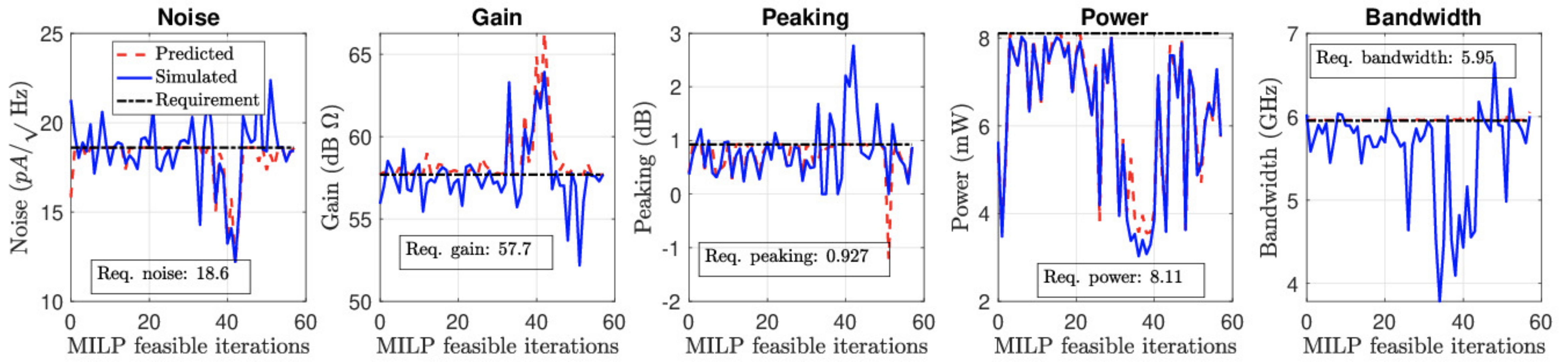}
    \centering
    \caption{Step 2 simulation results: Component selection for the GA-synthesized design (architecture+component)
with the aim of improving upon the human design. Results shown for iterations with a feasible solution 
to the MILP formulation.}
\label{fig:ArchBeatHuman}
\end{figure*}

\begin{figure*}[!ht]
    \includegraphics[scale=0.33]{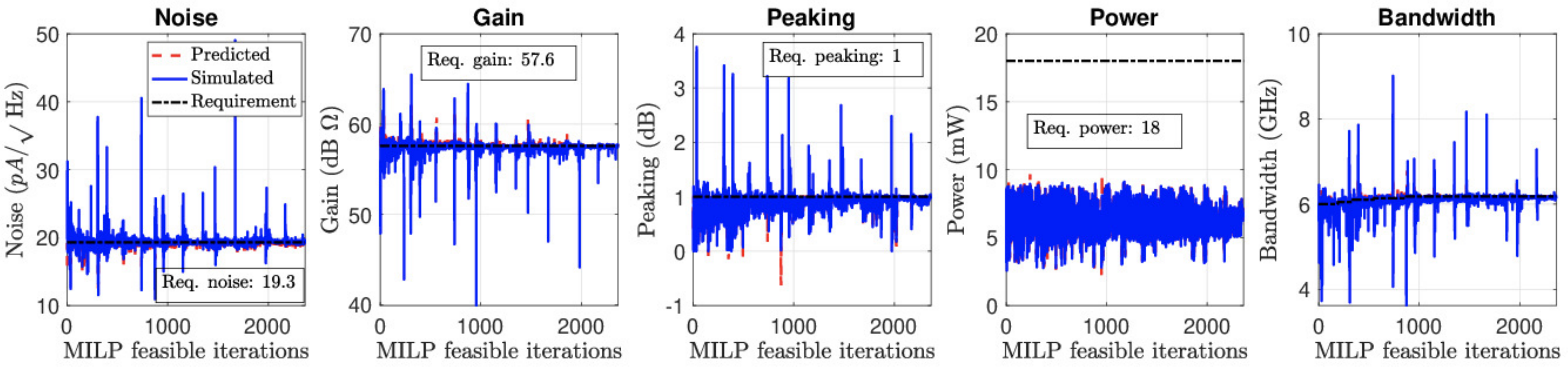}
    \centering
    \caption{Step 2 simulations results: Component selection for the GA-synthesized  design (architecture + component)
with the aim of maximizing bandwidth while meeting all the hard constraints.  The bandwidth 
requirement is increased gradually by $0.5\%$ on meeting the specified value in 
a trial.}
\label{fig:ArchMaxBW}
\end{figure*}

Fig.~\ref{fig:ArchBeatHuman} shows the simulation results for Step 2 during circuit synthesis 
with the aim of dominating the human design in all metrics. We show the values predicted by the NN 
in dotted-red, the result of applying that input and simulating it on the system in solid blue, and the 
requirement depicted by a black line (also shown in text). As the NN learns a better
system representation with the help of more training instances, the deviation of the value predicted 
by the NN from the actual simulation result reduces. A dominating circuit design is obtained in only 
311 more simulations (in addition to the 1600 simulations required in Step 1) within 2 CPU hours. 
Fig.~\ref{fig:ArchMaxBW} shows the simulation results for Step 2 during circuit synthesis to maximize 
bandwidth while meeting all the hard design constraints.
While maximizing the bandwidth, we encourage the method to 
perform better by increasing the bandwidth requirement that needs to be met by $0.5\%$ starting from 
6 GHz.  


\subsection{Component selection}
\label{subsec:TwoStageCompSelec}
Next, we evaluate DISPATCH when the architecture is fixed and component values need to
be selected. We illustrate it with the architecture for two-stage and  three-stage transimpedence 
amplifier architectures presented in \cite{wang2018learning}. 

\subsubsection{Two-stage transimpedence amplifier}
We use DISPATCH to determine the width of all the MOSFETs and the resistors with the goal of 
minimizing the objectives defined in Section \ref{subsec:ArchExample}.
We discretize the search space during GA-based component selection in Step 1 by generating 250 Sobol 
samples for resistors in the $[100, 5{\rm k}]\ \Omega$ range and width in the $[0.2, 50]\ \mu \rm m$ 
range.  These value ranges include those for the human-designed circuit. 
Table \ref{tab:GeneForCompSel} shows a chromosome for this case where $W_{i}$ corresponds to
the width of MOSFET $T_{i}$, and $R_{F}$ and $R_{6}$ are the values of the resistors in 
Fig.~\ref{fig:twoStageSeed}.

\begin{table}[htb]
\caption{Chromosome representation for component selection. The top row shows the parameter name for 
the component and the bottom row the value of that parameter.}
\label{tab:GeneForCompSel}
\resizebox{\columnwidth}{!}{%
\begin{tabular}{|c|c|c|c|c|c|c|c|}
\hline
$W_1$($\rm \mu m$) & $W_2$($\rm \mu m$) & $W_3$($\rm \mu m$) & $W_4$($\rm \mu m$) & $W_5$($\rm \mu m$) & $W_6$($\rm \mu m$) & $R_F$($\rm k\Omega$) & $R_6$($\rm k\Omega$)     \\ \hline
0.39 & 47.86 & 0.59 & 24.71 & 15.18 & 7.40 & 750.8 & 2798.8 \\ \hline
\end{tabular}
}
\end{table}

We evolve a generation of 30 individuals for a maximum of 400 generations. We use a smaller population 
size than in architecture search because the problem is simpler due to a smaller search space.  We 
use a tournament size of 10, mutation rate of 0.1, and crossover probability of 0.9: same as in 
architecture search.  The stopping criteria are:
\begin{itemize}
    \item The scaled bandwidth with respect to the human design falls below $1$. 
    \item The number of generations exceeds 100 and bandwidth stays the same for more than 100 
generations, indicating saturation.
    \item The maximum number of generations, i.e., 400, is reached.
\end{itemize}

\begin{figure*}[htbp]
\centering
\subfloat[]{\includegraphics[height=1.3in, width=2.in]{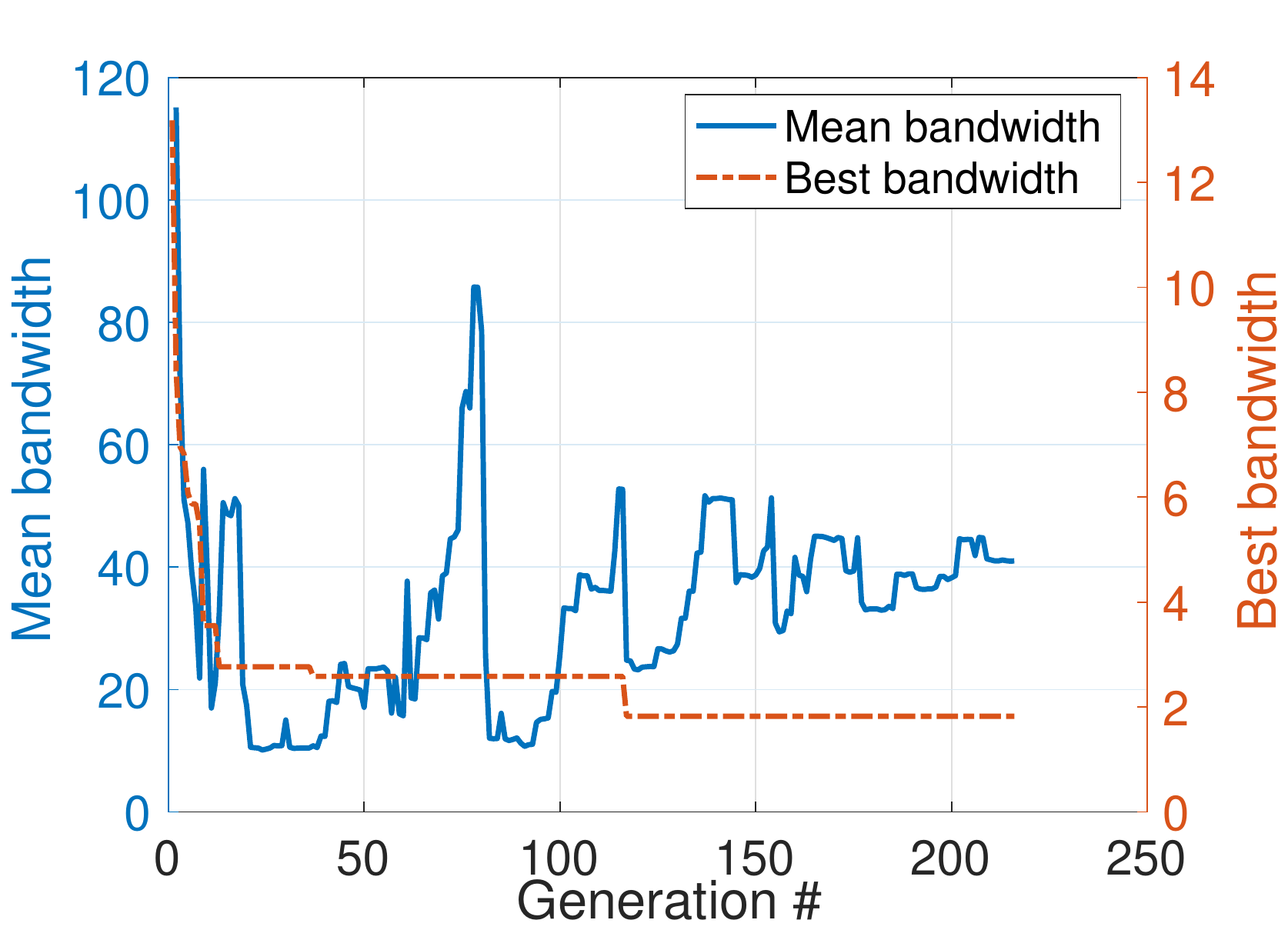}}
\subfloat[]{\includegraphics[height=1.3in, width=2.in]{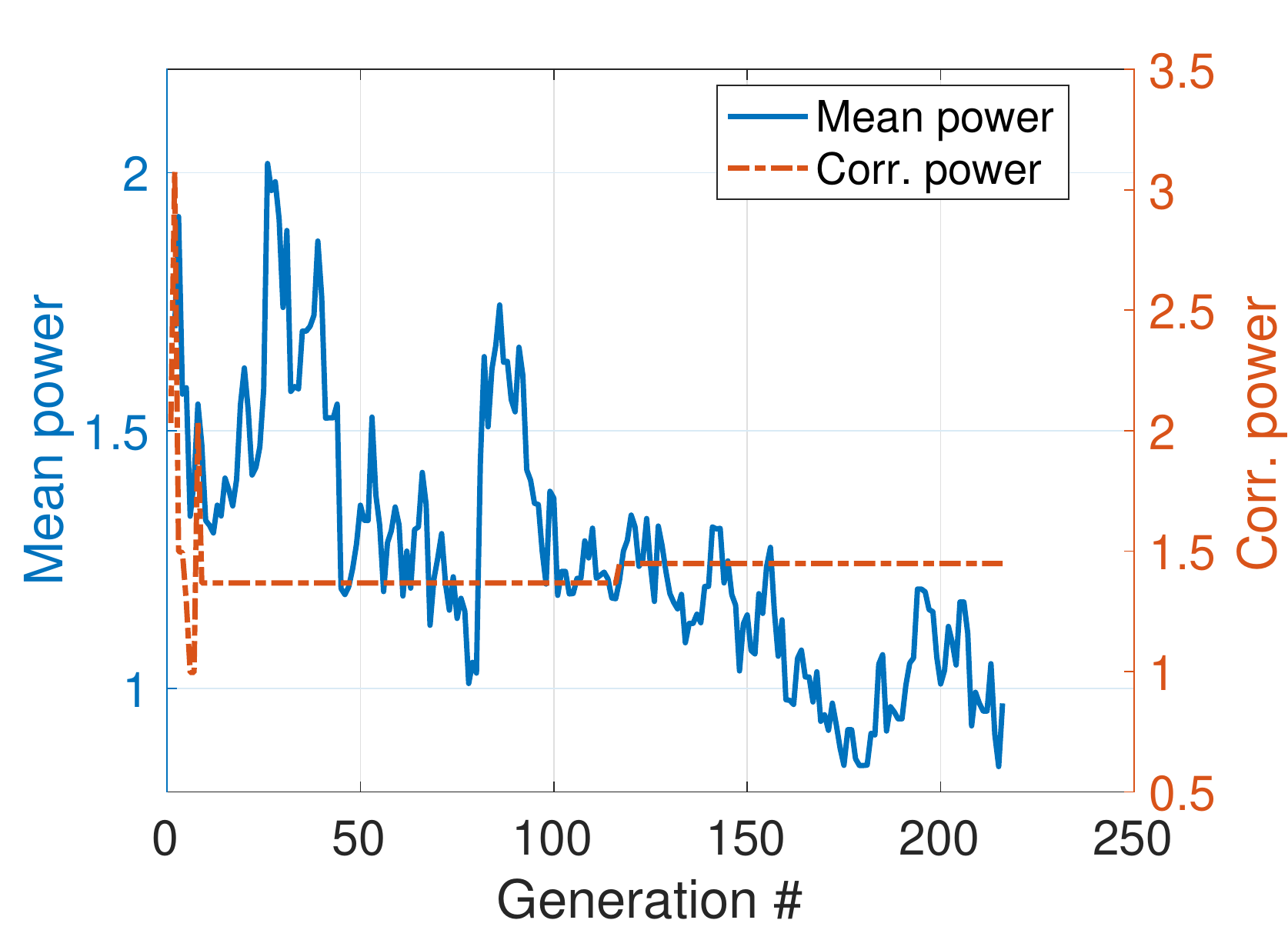}}
\subfloat[]{\includegraphics[height=1.3in, width=2.in]{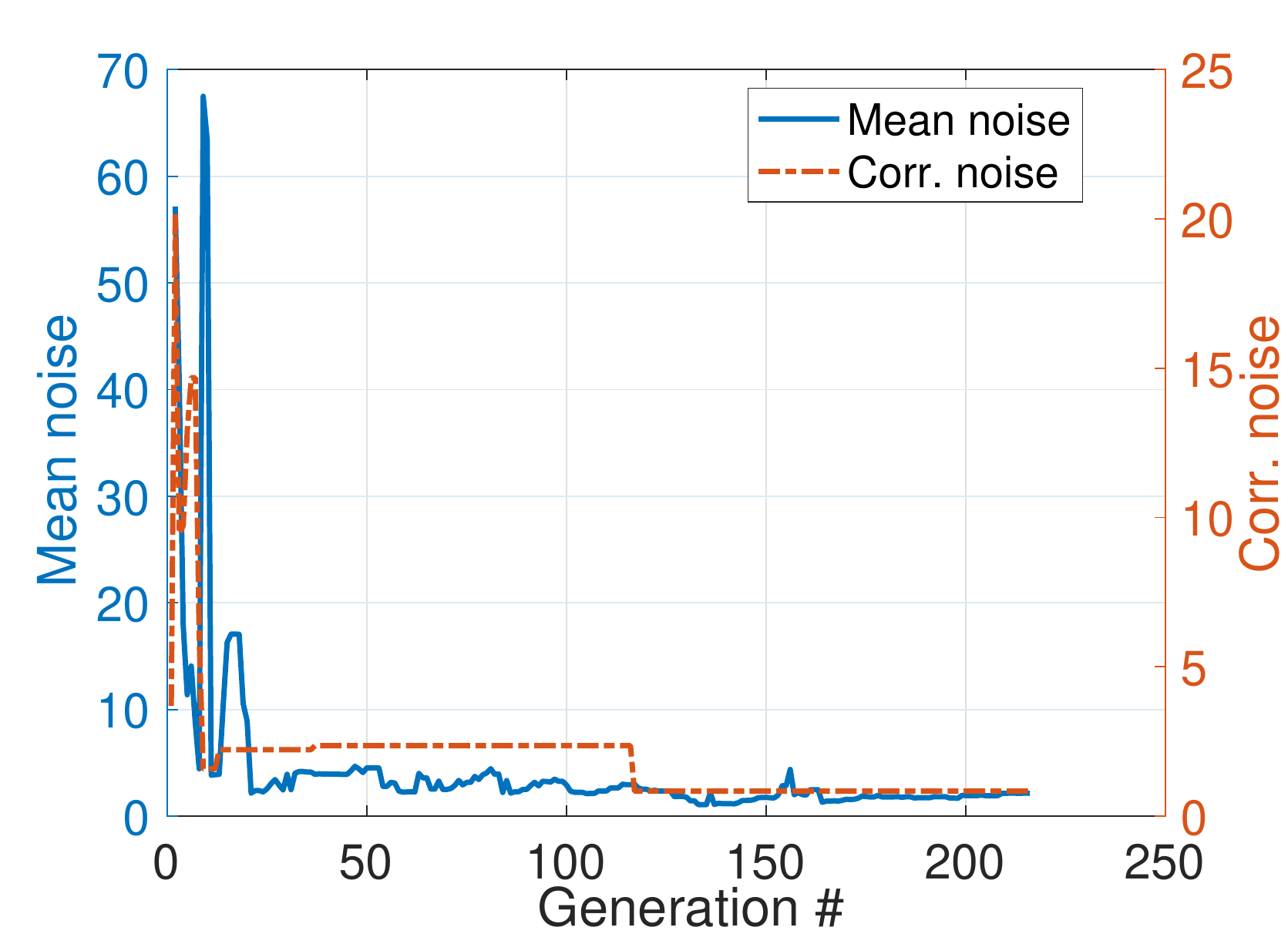}}
\caption{Scaled objectives across generation for (a) bandwidth, (b) power, and (c) noise from 
the second generation onwards for component selection of the two-stage 
transimpedence amplifier. The solid blue line shows the mean objectives averaged across 
all individuals in a generation and the dotted-orange line shows the individual with the best 
objective for (a) bandwidth and the corresponding (Corr.) objectives for (b) power, and (c) noise.}
\label{fig:GACompSelTwoStageEvolution}
\end{figure*}

Fig.~\ref{fig:GACompSelTwoStageEvolution} shows the evolution of objective values across GA
generations. The first few generations have high objective values as we do not use a seed design in 
component selection. GA stopped after the $216^{th}$ generation due to saturation in performance. 
Other observations are similar to those in Fig.~\ref{fig:GAAnalysisArch}. 
Synthesis required 0.9 CPU hour.  Table \ref{tab:GATwoStageComp} shows the values of
various objectives obtained after architecture synthesis. The bandwidth is inferior to human design although it is better in other metrics.
Hence, in Step 2, we fine-tune this design through component selection to improve its performance.

\begin{table*}[h]
\centering
\caption{Comparison of GA-synthesized circuit with the human-designed circuit for component selection 
for a two-stage transimpedence amplifier.}
\label{tab:GATwoStageComp}

\resizebox{\textwidth}{!}{\begin{tabular}{|c|c|l|c|c|c|c|c|c|}
\hline
 & \#Samples & Time & Noise ($\rm pA/\sqrt{Hz}$) & Gain ($\rm dB\  \Omega$) & Peaking ($\rm dB$) & Power ($\rm mW$) & Gate area (${\rm \mu m}^2$) & Bandwidth ($\rm GHz$) \\ \hline
Spec         & -       & -        & $\leq 19.3$ & $\geq 57.6$ & $\leq 1$ & $\leq 18$ & -       & maximize     \\ \hline
Human Design \cite{wang2018learning} & 1,289,618 & months   & 18.6         & 57.7         & 0.927    & 8.11      & 23.11 & 5.95  \\ \hline
GA           & 6,480    & 0.9 hr & 16.0         & 58.4         & 0.801    & 4.61      & 17.30   & 5.35 \\ \hline
\end{tabular}}
\end{table*}

\begin{table*}[h]
\centering
\caption{Comparison of designs synthesized with DISPATCH with those 
in \cite{wang2018learning} for the two-stage transimpedence amplifier (hard constraint
violation shown in a circle).} 
\label{tab:twoStCompSel}
\resizebox{\textwidth}{!}{\begin{tabular}{|c|c|l|c|c|c|c|c|c|}
\hline
 &
  \#Samples &
  Time &
  Noise ($\rm pA/\sqrt{\rm Hz}$) &
  Gain ($\rm dB\  \Omega$) &
  Peaking ($\rm dB$) &
  Power ($\rm mW$) &
  Gate area (${\rm \mu m}^2$) \footnote{\textit{Note: Gate area is shown only for designs for which this information was available. There was a problem in area calculation in \cite{wang2018learning} which was confirmed after contacting the authors.}} &
  Bandwidth ($\rm GHz$) \\ \hline
Spec          & -       &       & $\leq 19.3$ & $\geq 57.6$ & $\leq 1$ & $\leq 18$ & -       & maximize    \\ \hline
Human Design \cite{wang2018learning}  & 1,289,618 & months & 18.6         & 57.7         & 0.927    & 8.11      & 23.11 & 5.95 \\ \hline
{Human Design (DISPATCH)} &
  {7,168} &
  {7.8 hrs} &
  {18.4} &
  {57.8} &
  {0.870} &
  {5.52} &
  {16.85} &
  {5.95} \\ \hline
DDPG \cite{wang2018learning}          & 50,000   & 30 hrs & 19.2         & 58.1         & 0.963     & 3.18       & -       & 5.78 \\ \hline
{DDPG (DISPATCH)} &
  {9,285} &
  {125.5 hrs} &
  {19.2} &
  {58.3} &
  {0.943} &
  {3.18} &
  {10.88} &
  {5.75} \\ \hline
Bayesian Opt. \cite{wang2018learning} & 880     & 30 hrs & \enumber{19.6}         & 58.6         & 0.629     & 4.24       & -       & 5.16 \\ \hline
{Bayesian Opt. (DISPATCH)} &
  {6,661} & 
  {1.4 hrs} &
  {18.0} &
  {59.2} &
  {0.536} &
  {4.18} &
  {14.00} &
  {5.51} \\ \hline
{Max. Bandwidth (DISPATCH)} &
  {8,911} &
  {82.1 hrs} &
  {19.2} &
  {57.6} &
  {0.921} &
  {6.81} &
  {19.33} &
  {6.12} \\ \hline
\end{tabular}}
\end{table*}

\begin{figure*}[!ht]
    \includegraphics[scale = .32]{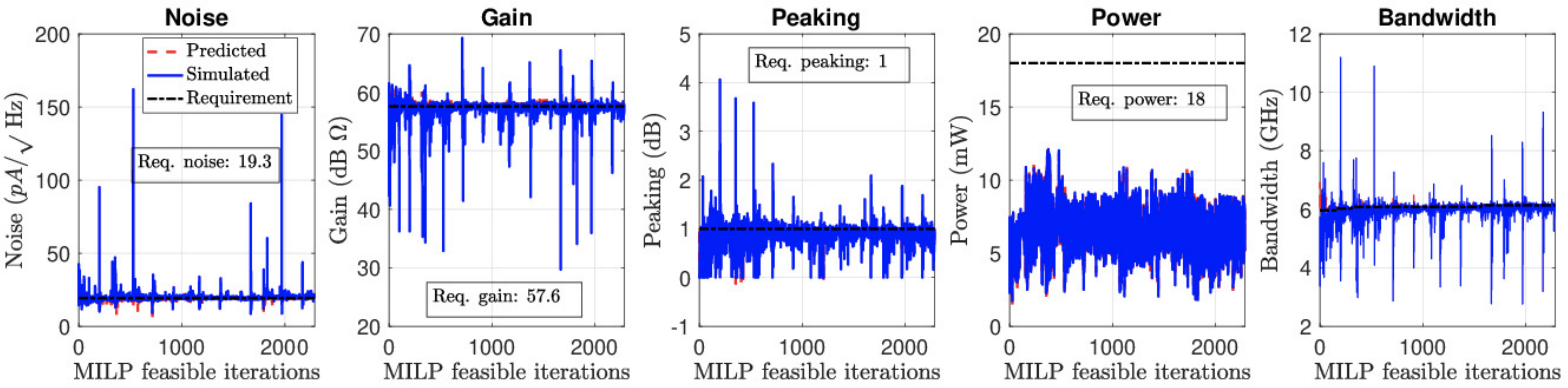}
    \centering
 \caption{Step 2 simulations results: Component selection for the GA-synthesized  (component only)
two-stage transimpedence amplifier with the aim of maximizing the bandwidth while meeting all the hard constraints.  The 
bandwidth requirement is increased gradually by $0.5\%$ on meeting the specified 
value in a trial.}
\label{fig:TwoStageCompMaxBW}
\end{figure*}

Table \ref{tab:twoStCompSel} shows comparisons of designs synthesized using DISPATCH with baseline designs. We set the required area to be the same as for architecture search. 
Our methodology performs
better across the board relative to the human design with only 8 CPU hours of effort and 
7168 simulations.  Our design has $0.5\%$ lower bandwidth than the DDPG design, but performs
better or same in terms of the other metrics.  Our design performs better than Bayesian optimization
based design across the board while also satisfying all the constraints (the former does not
satisfy the noise constraint).  Finally, the last row shows the design that meets all the hard 
constraints while maximizing bandwidth.  Fig. \ref{fig:TwoStageCompMaxBW} shows the simulation 
results for maximizing bandwidth while meeting all the hard constraints.
The bandwidth requirement is increased 
by $0.5\%$ starting from 5.96 GHz.

\subsubsection{Three-stage transimpedence amplifier}
\label{subsec:threestageCompSel}

\begin{figure}[!ht]
    \includegraphics[scale=0.23]{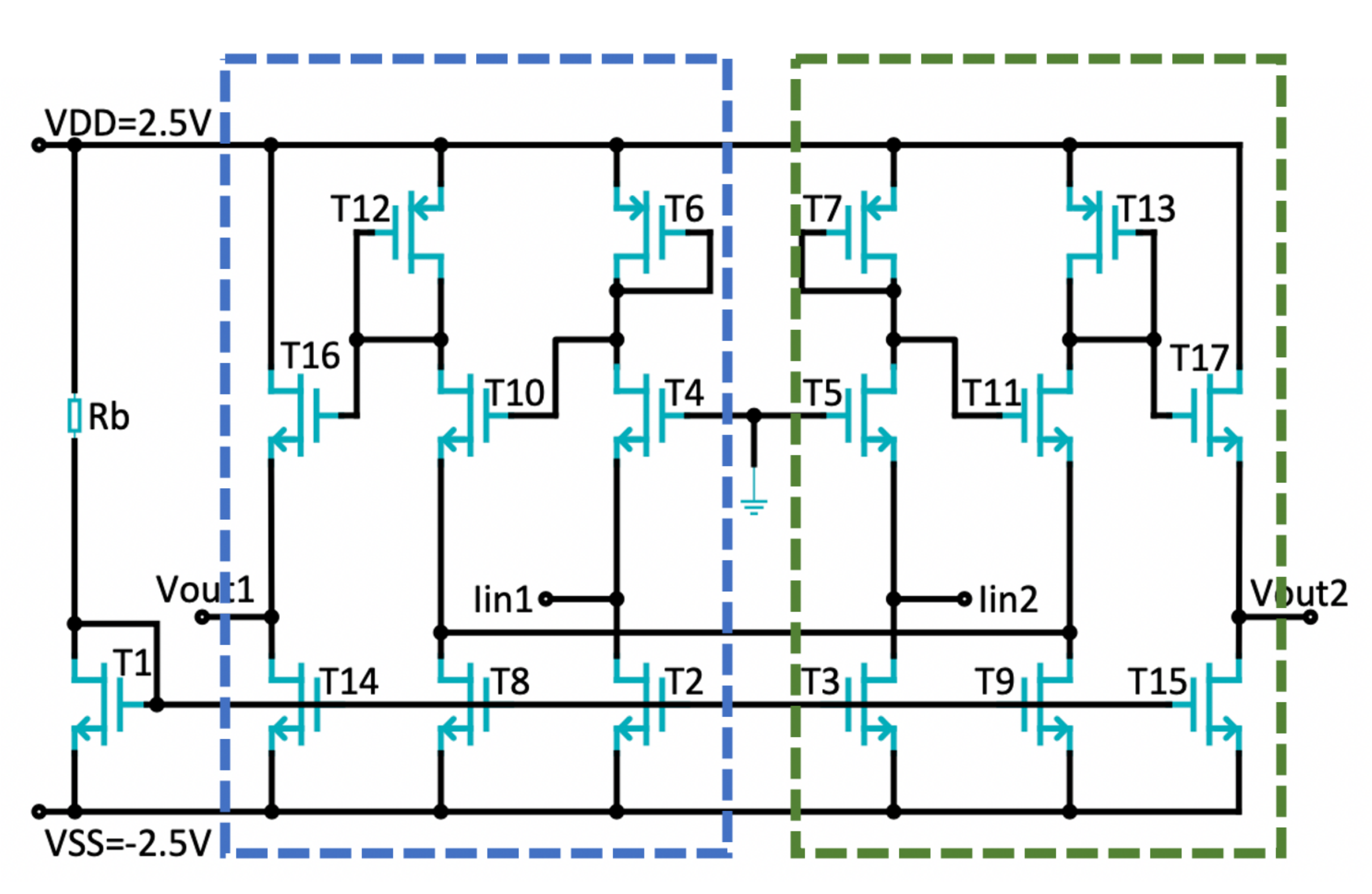}
    \centering
    \caption{Schematic of a three-stage differential transimpedence amplifier 
\cite{wang2018learning}.}
\label{fig:ThreeStageSchematic}
\end{figure}

Next, we discuss component selection for the three-stage transimpedence amplifier shown in 
Fig.~\ref{fig:ThreeStageSchematic}, which is adapted from \cite{wang2018learning}. The blue and green 
dotted boxes contain subcircuits that are mirror images of each other.  Hence, we obtain the
component values for only one subcircuit and mirror them in the other.  We determine the width 
and length of all the MOSFETs, including the bias transistor T1.  There are 19 components in
all: width/length of nine MOSFETs and a resistor \emph{Rb}.

The requirement is to minimize the sum of the gate areas of all the MOSFETs while meeting hard 
constraints for \textit{gain}, \textit{bandwidth}, and \textit{power}.  We discretize the search space 
during component selection in Step 1 by generating 250 Sobol samples with width in the 
$[2, 30]\ \mu \rm m$, length in the $[1, 2.2]\ \mu \rm m$, and resistor in the 
$[50{\rm k}, 500{\rm k}]\ \Omega$ range. The human-synthesized circuit has width in the 
$[2, 44]\ \mu \rm m$ range, length in the $[1, 2]\ \mu \rm m$ range, and a 
resistor of $291\ {\rm k}\Omega$.  We reduce the search space for width by around $30\%$ to encourage 
Step 1 to obtain designs with a smaller area. We also round the length  to the nearest $0.2$ units 
since the technology only permits the length to be an integer multiple of $0.2\ \mu \rm m$. We use 
this rounding in Step 2 as well.

We use the same parameters for GA (\#generations, population size, tournament size, mutation rate, 
crossover probability, stopping criteria) as used for component selection for the two-stage 
transimpedence amplifier. We have the following objectives:

\begin{enumerate}
    \item \textit{Bandwidth}: We set the passband to $90~{\rm MHz}$ with a target gain of 
$35.57\  {\rm k}\Omega$ (this corresponds to a gain of $85\  {\rm dB}$ for the half-circuit whereas the minimum 
gain required for its valid design is $80\  {\rm dB}$ or $20\  {\rm k}\Omega$ differential gain). We use rewards/penalties as follows.
    \begin{itemize}
        \item A reward for a MOSFET operating in the saturation region, else a penalty is applied as 
follows:
        \begin{itemize}
            \item \textit{Saturation region}: Reward of 1 divided by the number of MOSFETs. 
            \item \textit{Linear region}: Penalty of 5 divided by the number of MOSFETs. 
            \item \textit{Cutoff region}: Penalty of 7 divided by the number of MOSFETs.
        \end{itemize}
    Since the number of MOSFETs is larger than in the earlier design, we slightly increase these 
numbers to encourage MOSFET operation in the saturation region.
    \item Penalty of 15 based on fractional deviation in gain below 80 dB ($20\ {\rm k\Omega}$) for the half circuit in Fig. \ref{fig:ThreeStageSchematic}. 
        \item Penalty of 15 based on fractional deviation in bandwidth below $90~ {\rm MHz}$.
    \end{itemize}
    \item \textit{Area}: We define the area objective as the ratio of the sum of MOSFET areas in one 
of the mirrored regions and the area of the bias MOSFET (T1) and the corresponding areas in the 
human-synthesized circuit. 
    \item \textit{Power}: We define this objective as the ratio between measured power and the power consumed by human designed circuit which is $1.37~\rm mW$.
We levy a penalty of 15 based on the fractional deviation in power 
above this value. 
\end{enumerate}

\begin{figure*}[htbp]
\centering
\subfloat[]{\includegraphics[height=1.3in, width=2.in]{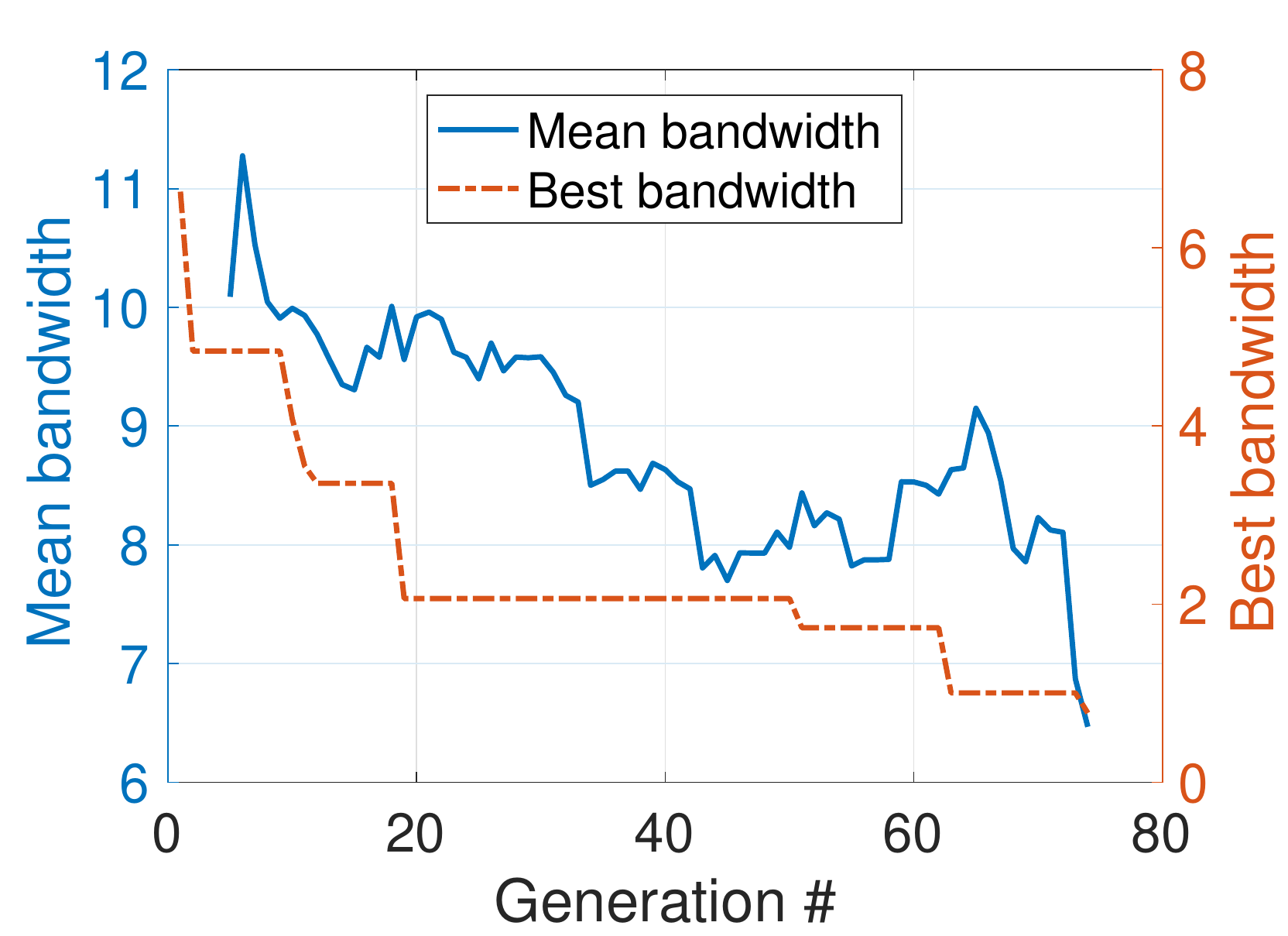}}
\subfloat[]{\includegraphics[height=1.3in, width=2.in]{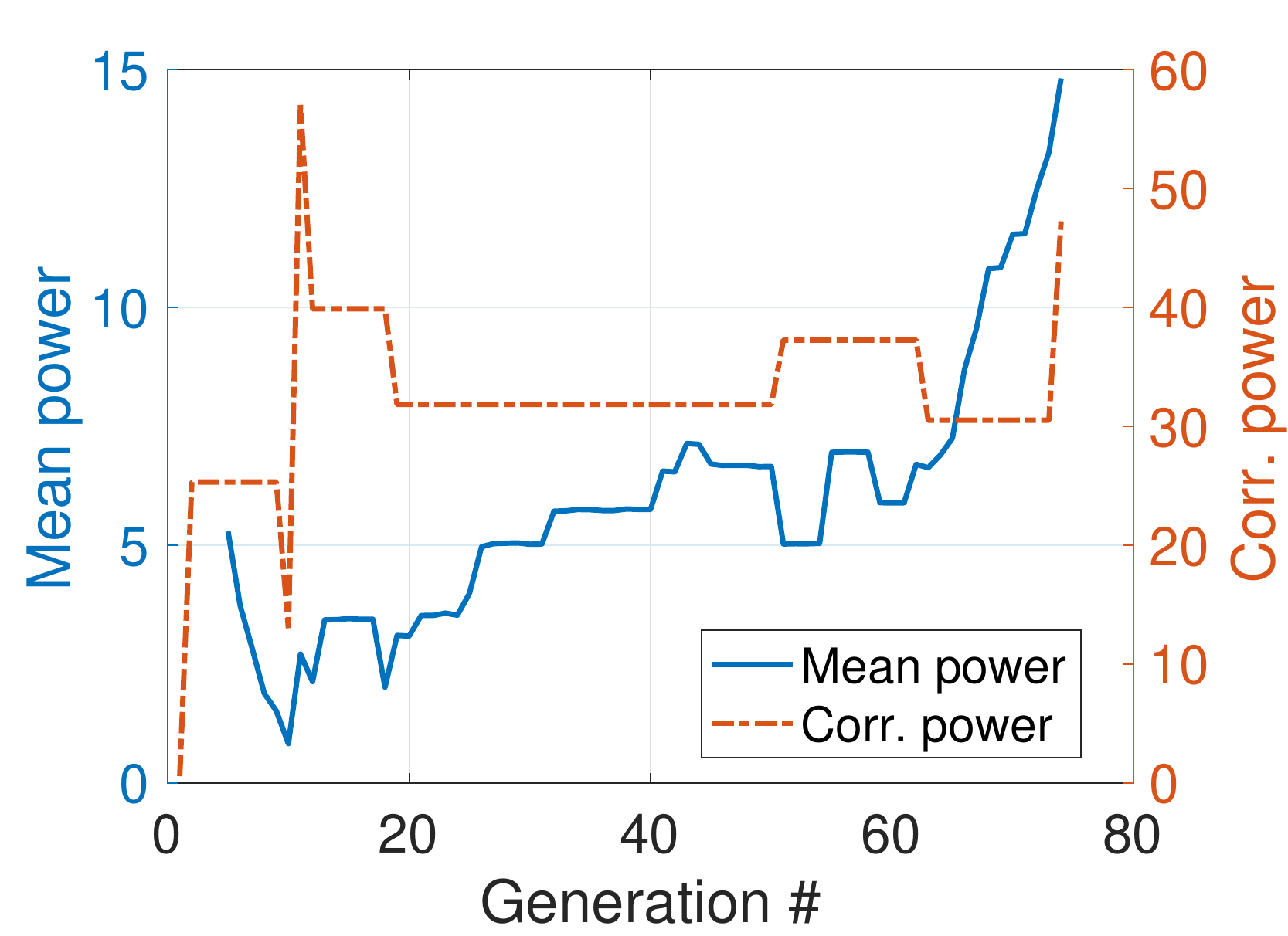}}
\subfloat[]{\includegraphics[height=1.3in, width=2.in]{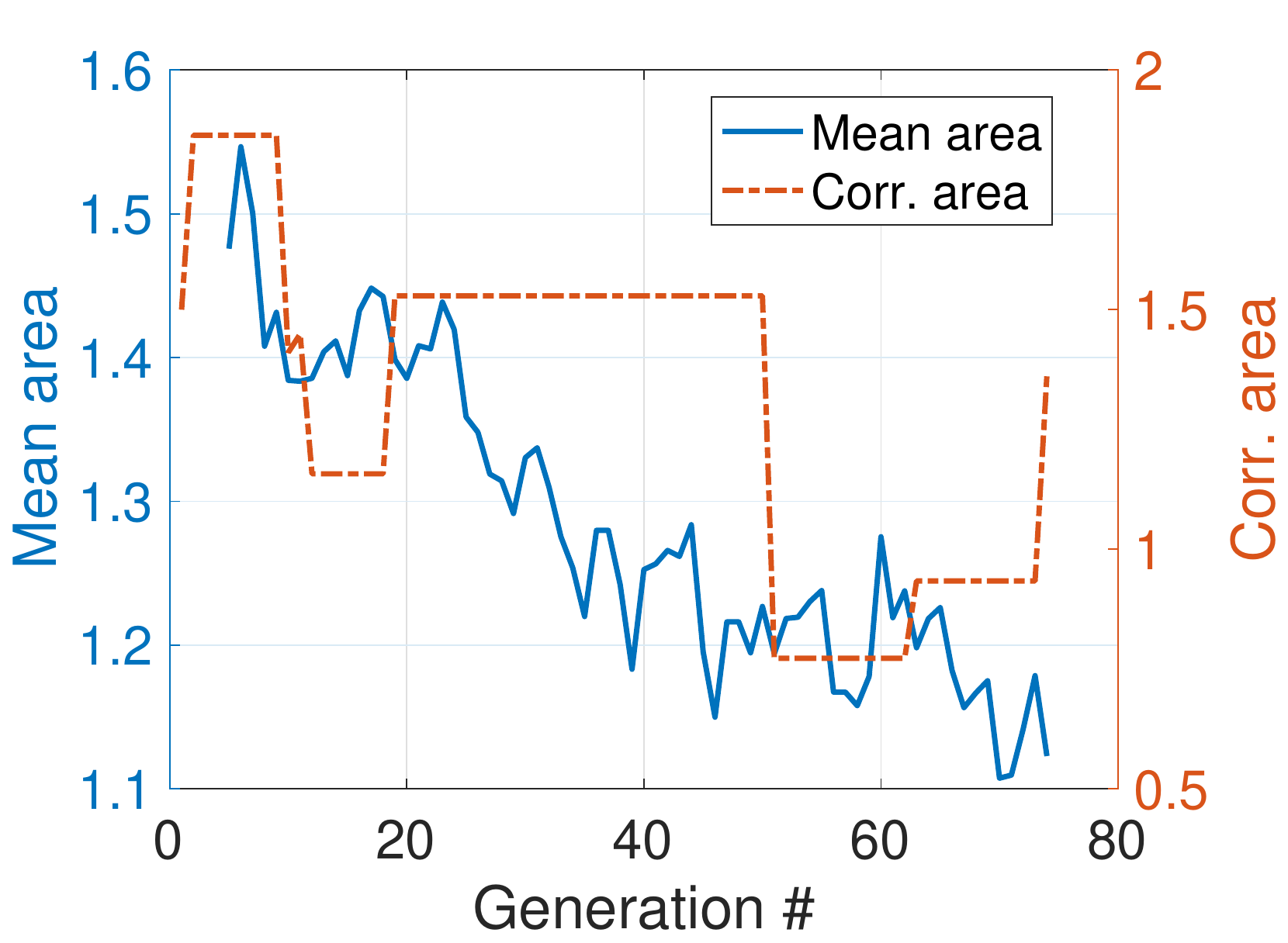}}
\caption{Scaled objectives across generation for (a) bandwidth, (b) power, and (c) area from
the fifth generation onwards for component selection of the three-stage 
transimpedence amplifier. The solid blue line shows the mean objectives averaged across 
all individuals in a generation and the dotted-orange line shows the individual with the best 
objective for (a) bandwidth and the corresponding (Corr.) objectives for (b) power, and (c) area.}
\label{fig:GAAnalysisThreeStageComp}
 
\end{figure*}

Fig.~\ref{fig:GAAnalysisThreeStageComp} shows the evolution across GA generations.
Mean objectives across all individuals in a generation after the fifth generation for bandwidth, 
power, and area are shown in blue. There is a trade-off among the three objectives. These observations 
are similar to those in Fig.~\ref{fig:GACompSelTwoStageEvolution}.
GA is terminated after 74 generations as the objective for bandwidth is reduced to less than 1. Table \ref{tab:GAThreeStageCompSelectionComp} shows a comparison of the circuit synthesized using GA with human design. The circuit obtained by GA violates the hard constraints on bandwidth and power and is hence not an acceptable design. The area of this circuit is much higher than the human-designed circuit.

\begin{table}[h]
\centering
\caption{Comparison of GA-synthesized circuit with the human-designed circuit for component selection 
for a three-stage transimpedence amplifier (hard constraint violation shown in a circle).}
\label{tab:GAThreeStageCompSelectionComp}
\resizebox{\columnwidth}{!}{%
\begin{tabular}{|c|c|l|c|c|c|c|}
\hline
            & \#Samples     & Time              & Bandwidth ($\rm MHz$) & Gain ($\rm k \Omega$) & Power ($\rm mW$)    & Gate area (${\rm \mu m}^2$) \\ \hline
Spec         & -       & -      & $\geq 90$ & $\geq 20$ & $\leq 3$ & -   \\ \hline
Human Design \cite{wang2018learning} & 10,000,000 & months & 90.1       & 20.2       & 1.37      & 211 \\ \hline
{GA} & {2,220} & {0.31 hr} & {\enumber{86.7}}    & {30.5}               & {\enumber{5.33}} & {289.5}     \\ \hline
\end{tabular}
}
\end{table}

\begin{figure*}[!ht]
    \includegraphics[scale = .32]{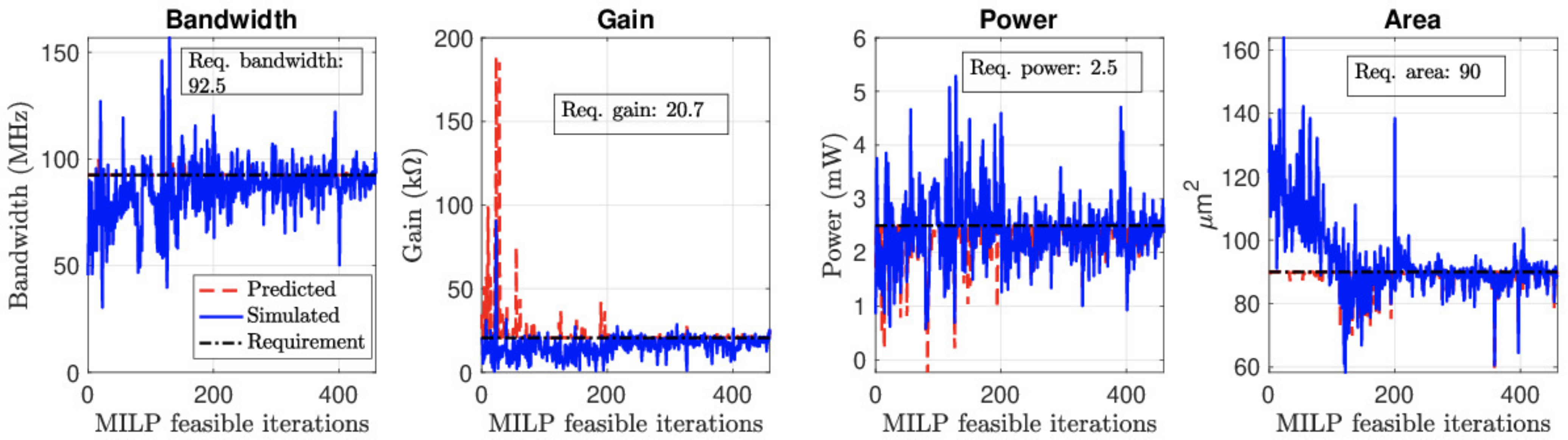}
    \centering
    \caption{Step 2 simulation results: Component selection for the GA-synthesized (component only) three-stage transimpedence amplifier  with the aim of improving upon the DDPG design. Results shown for iterations with a feasible solution to the MILP 
formulation.}
\label{fig:ThStgCNMADominateDDPGV1}
\end{figure*}

In the next step, we fine-tune the component values to  obtain designs that dominate the benchmark 
designs from \cite{wang2018learning} or perform close to it.  Besides, we also synthesize a circuit 
that meets all the hard constraints while minimizing the gate area. 
Table \ref{tab:ThreeStageCompSelCNMA} shows a comparison of designs obtained using DISPATCH with other
designs. DISPATCH synthesizes a design that performs close to human design in 129 CPU hours, using 
5025 simulations, while meeting all the hard constraints. The circuit consumes $0.05\  \rm mW$ more 
power compared to the human design but is better in all other metrics. A dominating design compared to 
DDPG is obtained in 20 CPU hours with about $14\times$ fewer samples. 
Fig.~\ref{fig:ThStgCNMADominateDDPGV1} shows the simulation results for Step 2 during circuit 
synthesis for this case. The dominating design in comparison to Bayesian optimization based design 
is attained in less than 1 CPU hour of total simulation time. DISPATCH synthesizes a design that has 
the least area compared to all other designs with a reduction of about $25\%$ from the best design 
achieved by DDPG. The initial requirement of area of $80\ \rm {\mu m}^2$ is gradually reduced by $5\%$ as 
soon as the requirement is met.

\begin{table*}[]
\caption{Comparison of designs synthesized using DISPATCH with those synthesized
in \cite{wang2018learning} for the three-stage transimpedence amplifier (hard constraint
violations shown in a circle).} 

\label{tab:ThreeStageCompSelCNMA}
\centering
\begin{tabular}{|c|c|c|c|c|c|c|}
\hline
\multicolumn{1}{|c|}{} &
  \multicolumn{1}{c|}{\#Samples} &
  Time &
  \multicolumn{1}{c|}{Bandwidth ($\rm MHz$)} &
  \multicolumn{1}{c|}{Gain ($\rm k \Omega$)} &
  \multicolumn{1}{c|}{Power ($\rm mW$)} &
  \multicolumn{1}{c|}{Gate area (${\rm \mu m}^2$)} \\ \hline
\multicolumn{1}{|c|}{Spec} &
  \multicolumn{1}{c|}{-} &
  - &
  \multicolumn{1}{c|}{$\geq 90$} &
  \multicolumn{1}{c|}{$\geq 20$} &
  \multicolumn{1}{c|}{$\leq 3$} &
  \multicolumn{1}{c|}{-} \\ \hline
\multicolumn{1}{|c|}{Human Design \cite{wang2018learning}} &
  \multicolumn{1}{c|}{10,000,000} &
  months &
  \multicolumn{1}{c|}{90.1} &
  \multicolumn{1}{c|}{20.2} &
  \multicolumn{1}{c|}{1.37} &
  \multicolumn{1}{c|}{211.0} \\ \hline
\multicolumn{1}{|c|}{{Human Design (DISPATCH)}} &
  \multicolumn{1}{c|}{{5,025}} &
  {128.5 hrs} &
  \multicolumn{1}{c|}{91.0} &
  \multicolumn{1}{c|}{{20.3}} &
  \multicolumn{1}{c|}{{1.42}} &
  \multicolumn{1}{c|}{{174.7}} \\ \hline
DDPG \cite{wang2018learning} &
  40,000 &
  40 hrs &
  92.5 &
  20.7 &
  2.50 &
  90.0 \\ \hline
{DDPG (DISPATCH)} &
  {2,869} &
  {19.9 hrs} &
  {93.2} &
  {25.9} &
  {2.37} &
  {88.6} \\ \hline
Bayesian Opt. \cite{wang2018learning} &
  1,160 &
  40 hrs &
  \enumber{72.5} &
  21.1 &
  \enumber{4.25} &
  130.0 \\ \hline
{Bayesian Opt.(DISPATCH)} &
  {2,334} &
  {0.4 hr} &
  {99.4} &
  {25.7} &
  {3.00} &
  {120.5} \\ \hline
{Min. Area (DISPATCH)} &
  {4,695} &
  {124.1 hrs} &
  {90.0} &
  {20.7} &
  {2.97} &
  {67.2} \\ \hline
\end{tabular}
\end{table*}

\section{Conclusion}
\label{sec:conclusion}
We formulated CPS design as an MOO problem and proposed DISPATCH, a two-step CPS synthesis 
methodology, for solving it. We used gradient-free search through GA for exploration and an NN to 
enable sample efficiency by formulating it as an MILP problem to generate samples based on 
\emph{active learning}. The methodology provides a flexible framework for discovering efficient 
architectures and component values or simply the component values for a fixed architecture. Using 
this framework, we were able to improve the sample efficiency by a factor of 5-14$\times$ compared 
to the most sample-efficient designs synthesized by DDPG. 

As part of future work, we plan to represent CPS as graphs to explore different architectures 
through manipulation of these graphs and explore other sample-efficient techniques for 
\emph{active learning}.  We also plan to enhance the efficacy of the methodology by combining
it with RL.

\vspace*{2mm}
\noindent
{\bf Acknowledgment:}
We would like to thank Hanrui Wang for providing details of his work and the simulation environment 
used in \cite{wang2018learning}.  The simulations presented in this article were performed on 
computational resources managed and supported by Princeton Research Computing, a consortium of groups 
including the Princeton Institute for Computational Science and Engineering (PICSciE) and the Office 
of Information Technology's High Performance Computing Center and Visualization Laboratory at 
Princeton University.

\bibliographystyle{IEEEtran}

\end{document}